\documentclass{omniareport}

\usepackage[toc,page,header]{appendix}

\usepackage{minitoc}
\usepackage{amsmath,amssymb}
\usepackage{booktabs}
\usepackage{multirow}
\usepackage{array}
\usepackage{colortbl}
\usepackage{subcaption}

\usepackage{graphicx}
\usepackage[most]{tcolorbox}
\tcbuselibrary{listings,breakable,skins}
\usepackage{listings}
\usepackage{tabularx}
\usepackage{adjustbox}
\usepackage{wrapfig}

\usepackage{makecell}
\usepackage{pifont}

\providecommand{\cmark}{\textcolor{green!55!black}{\ding{51}}}
\providecommand{\xmark}{\textcolor{red!70!black}{\ding{55}}}
\providecommand{\pmark}{}
\renewcommand{\pmark}{\textcolor{RoyalBlue}{$\blacktriangle$}}
\newcommand{\VerifyCode}{\textit{VerifyCode}}

\definecolor{HeaderGray}{RGB}{242,244,247}
\definecolor{GroupGray}{RGB}{232,235,240}
\definecolor{FoundationBlue}{RGB}{233,242,255}
\definecolor{ScaleGreen}{RGB}{235,248,240}
\definecolor{AgentOrange}{RGB}{255,243,230}
\definecolor{RuleGray}{RGB}{180,185,192}

\definecolor{RoleUser}{RGB}{54,88,153}
\definecolor{RoleAssistant}{RGB}{46,125,93}
\definecolor{RoleTool}{RGB}{161,98,7}
\definecolor{ActionBlue}{RGB}{33,101,181}

\definecolor{HeaderBlue}{HTML}{EAF1FB}
\definecolor{OverallGray}{HTML}{F2F4F7}

\definecolor{DemoUserBg}{RGB}{245,248,255}
\definecolor{DemoUserFrame}{RGB}{86,116,185}

\definecolor{DemoTraceBg}{RGB}{246,250,247}
\definecolor{DemoTraceFrame}{RGB}{78,145,108}

\definecolor{DemoAgentBg}{RGB}{244,251,247}
\definecolor{DemoToolBg}{RGB}{255,248,238}
\definecolor{DemoMsgFrame}{RGB}{205,211,220}
\definecolor{DemoRubricBg}{RGB}{255,247,248}
\definecolor{DemoRubricFrame}{RGB}{190,92,108}
\definecolor{RoleRubric}{RGB}{153,54,70}
\definecolor{RubricRed}{RGB}{252,232,235}

\definecolor{MutedText}{RGB}{95,102,115}

\title{OmniaBench: Benchmarking General AI Agents Across Diverse Scenarios}

\affiliation[]{%
  \shortstack[c]{%
    \textbf{Huawei Cloud Post-Training Team} \\
    \textbf{PKU DCAI Team} \\
    (Full author list in Contributions)
  }
}

\abstract{
Large language models are increasingly evolving from text generators into general agents capable of understanding user requests, invoking external tools, and completing complex tasks through interaction. However, existing agent benchmarks often focus on limited scenarios, tool ecosystems, or interaction formats, making it difficult to systematically characterize model capabilities across heterogeneous application settings. We introduce \textbf{OmniaBench}, a benchmark for evaluating general agents across diverse scenarios with explicit state spaces. We derive application-oriented scenario knowledge from app stores, product documents, industry resources, Web retrieval, and human refinement, forming a hierarchical taxonomy that spans ToC, ToB and ToE with 90 level-1 and 354 level-2 domains. Based on this taxonomy, we construct executable environments and synthesize single-turn and multi-turn tasks through four complementary routes: DAG, DAG-S, Solver, and Program. OmniaBench further introduces a ten-dimensional capability taxonomy and eight compositional atomic difficulty factors to support fine-grained evaluation and analysis. The resulting dataset contains 1,431 tasks, together with a challenging subset of 644 tasks designed to reduce evaluation cost and mitigate potential contamination of the full set after public release. The bench presents substantial challenges to current frontier models, with even Claude-Sonnet-5 and GPT-5.6-Sol achieving Overall Pass@1 scores of only 58.54 and 57.14, respectively. Further analyses reveal clear differences across domains and capabilities, as well as persistent limitations in planning, constraint maintenance, and adaptive correction. OmniaBench provides a broad and diagnostic benchmark for characterizing the capability boundaries of general agents.
}

\projectpage{\url{https://scuuy.github.io/OmniaBench}}

\code{\url{https://github.com/scuuy/OmniaBench}}

\begin{document}
\maketitle


\begin{figure*}[h]
    \vspace{-2em}
    \centering
    \includegraphics[width=\textwidth]{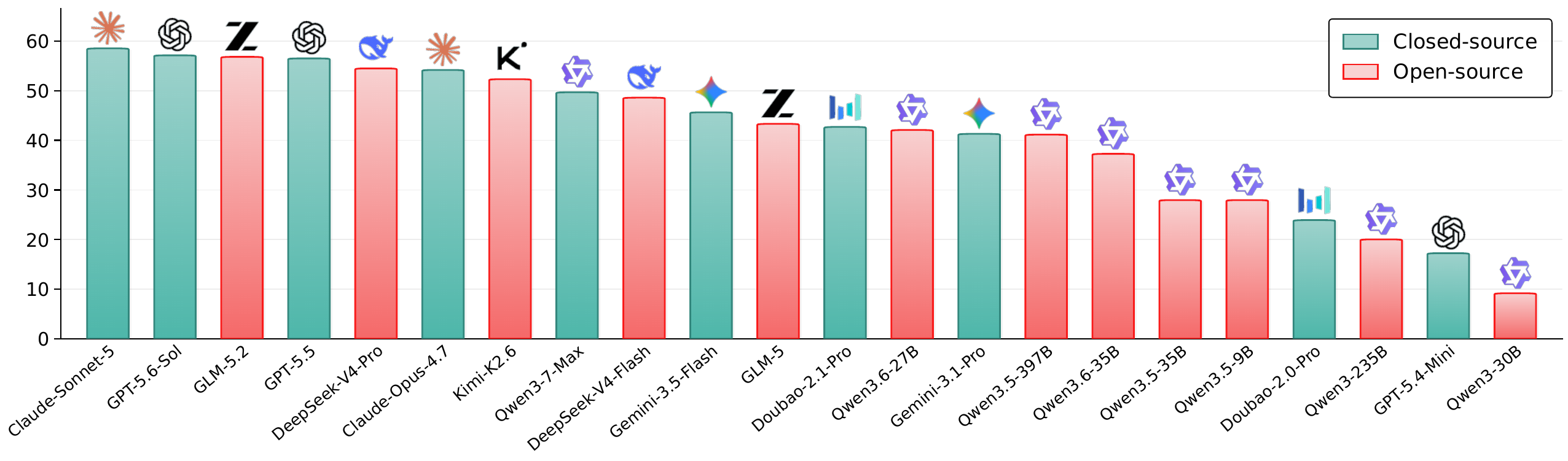}
    \caption{\textbf{OmniaBench leaderboard.} Overall performances of evaluated models on the challenging set.}
    \label{fig:leaderboard}
\end{figure*} 

\section{Introduction}

Recent advances in large language models (LLMs) have substantially improved their capabilities in language understanding, complex reasoning, code generation, and multi-turn interaction~\cite{deepseekai2026deepseekv4highlyefficientmilliontoken, glm5team2026glm5vibecodingagentic, kimiteam2026kimik25visualagentic}. These advances are gradually shifting LLMs from static text generators toward agentic systems capable of executing real-world tasks~\cite{liang2025dataflowllmdrivenframeworkunified, dong2026agent}. Enabled by function calling, retrieval augmentation, code execution, and coordinated tool use, LLM-based agents have shown strong potential in software development, information seeking, office automation, business workflows, and personal assistance, attracting increasing attention from both academia and industry~\cite{yao2024tau, boisvert2024workarena++, yang2024swe, jimenez2024swe}. Compared with conventional question answering or isolated reasoning tasks~\cite{cobbe2021training, wang2024mmlu}, such agents must operate in more open-ended settings, where they need to understand user intents, formulate multi-step plans, invoke external tools, maintain task states, and progressively complete user goals through interaction.

Around LLM-based agent evaluation, prior work has made substantial progress along several directions. Function-calling and tool-use benchmarks evaluate whether models can select appropriate tools, generate valid arguments, compose multiple tool calls, and satisfy execution constraints~\citep{patil2025berkeley, chen2025acebench, li2025tool, shen2026one}. Interactive benchmarks further introduce simulated users, dynamic states, and domain-specific rules, allowing agents to complete tasks through multi-turn interaction with external environments~\citep{yao2024tau, barres2025tau, he2025vitabench}. More recent studies have expanded evaluation toward open-ended and realistic settings, such as long-horizon planning, professional workflows, web and application environments, multimodal evidence grounding, and standardized tool ecosystems~\citep{froger2026gaia2, zhang2026deepplanning, patwardhan2025gdpval, backlund2025vending, bandi2026mcp, meng2026clawmark}. Despite these advances, existing benchmarks are often organized around a limited set of scenarios, tool ecosystems, or task formats, leaving a gap between benchmark coverage and the diversity of real-world agent use cases. In practice, general agents are expected to operate across heterogeneous applications and user needs, while handling multi-turn interaction, evolving task states, tool execution, and user-specific constraints. This motivates the need for a general-agent benchmark with broader scenario coverage, richer task forms, and more diagnostic evaluation dimensions.

\begin{figure*}[t]
    \centering
    \includegraphics[width=\textwidth]{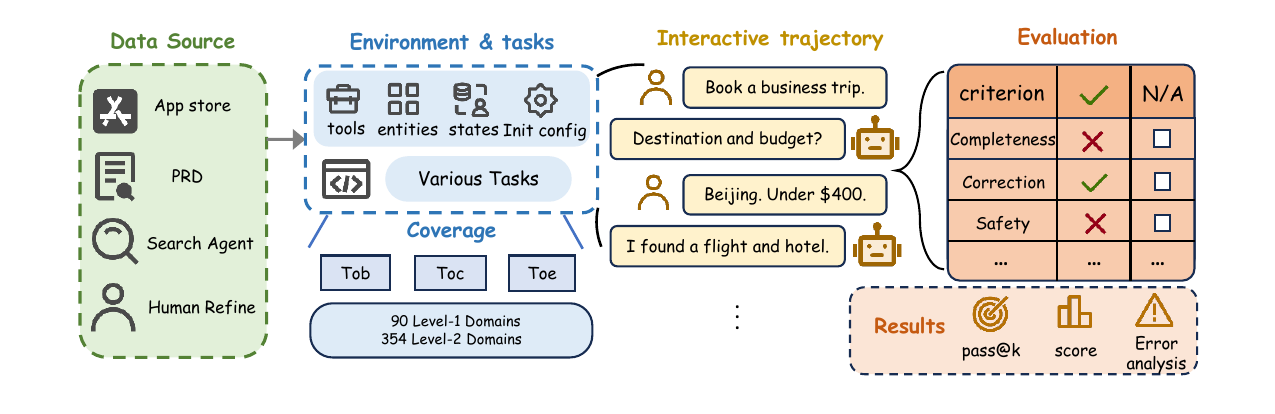}
    \caption{\textbf{Conceptual overview of OmniaBench.} The figure summarizes the key components of the benchmark, including data sources, executable task environments, interactive agent trajectories, and evaluation protocols. OmniaBench is designed to cover diverse ToC, ToB and ToE domains and to evaluate general agents with both rubric\&code based criteria and aggregate performance metrics.}
    \label{fig:benchmark_pipeline}
    \vspace{-1.2em}
\end{figure*}

To bridge this gap, we introduce \textbf{OmniaBench}, a multi-scenario and interactive benchmark for evaluating general agents. OmniaBench aims to construct agent tasks from real-world application ecosystems, covering a broad spectrum of user needs and application workflows. As illustrated in Figure~\ref{fig:benchmark_pipeline}, we collect and organize scenario knowledge from popular app stores, product requirement documents (PRD), web-search agents, and human refinement, resulting in a full scenario taxonomy that groups real-world agent applications into ToC, ToB and ToE settings. Built upon this taxonomy, OmniaBench covers 90 level-1 domains and 354 level-2 domains, providing a substantially broader domain space than benchmarks centered on a single tool ecosystem or a small set of task environments. We further instantiate these domains into interactive environments with tools, entities, states, and initialization configurations, and construct user requests across diverse task forms. As a result, models are required to understand user goals, invoke tools, track evolving states, and complete tasks through interaction.

OmniaBench further incorporates structured designs for capability measurement, task difficulty, and diagnostic evaluation. 
We define a ten-dimensional capability taxonomy covering Task Understanding, Information Gathering, Planning \& Decision Making, State Management, Tool Use, Code \& Programmatic Operations, Data Analysis, Office \& Document Handling, Interactive Collaboration, Reliability \& Safety.
This taxonomy allows evaluation results to be decomposed beyond overall success rates, exposing model strengths and weaknesses across distinct stages of agent execution. We also introduce a compositional atomic difficulty design, which organizes task challenges around user intent, persona constraints, environment states, tool execution, multi-turn interaction, and result verification. Together with rubric checklists and VerifyCode-based assessment, OmniaBench supports fine-grained analyses over domain categories, capability dimensions, task complexity, interaction turns, and error patterns, offering a systematic view of the capability boundaries of current general agents.

Our contributions are summarized as follows:

\begin{itemize}
    \item We introduce \textbf{OmniaBench}, a scenario-rich and challenging benchmark for evaluating general agents. Grounded in real-world application ecosystems, OmniaBench integrates popular app stores, PRD, web search, and human refinement to construct a broad scenario taxonomy, which is further instantiated into interactive and evaluable agent tasks. Under our evaluation protocol, even the frontier model Claude-Sonnet-5 and GPT-5.6-Sol achieves only 58.54 and 57.14 overall scores, respectively, demonstrating that OmniaBench poses substantial challenges to current frontier models.

    \item We develop a multi-dimensional \textbf{capability} taxonomy for
    general-agent evaluation. Instead of organizing tasks solely by application domains, OmniaBench characterizes model capabilities along ten dimensions: Task Understanding, Information Gathering, Planning \& Decision Making, State Management, Tool Use, Code \& Programmatic Operations, Data Analysis, Office \& Document Handling, Interactive Collaboration, and Reliability \& Safety, enabling fine-grained diagnostic analysis.

    \item We design a compositional \textbf{atomic difficulty} framework to characterize fine-grained challenges in general-agent tasks. This framework covers user intent and persona constraints, goal decomposition, environment state changes, tool execution, multi-turn interaction, error recovery, and result verification, allowing OmniaBench to construct tasks with diverse execution difficulty and interaction complexity.

    \item We conduct a systematic evaluation and analysis of both closed and open source models. Our experiments examine model performance from multiple perspectives, including overall results, domain categories, capability dimensions, task complexity, interaction turns, error distributions, and evaluation stability, providing detailed evidence for understanding the capability boundaries of current general agents.
\end{itemize}

\section{Related Work}

\begin{table*}[htbp]
\centering
\footnotesize
\setlength{\tabcolsep}{3.0pt}
\renewcommand{\arraystretch}{1.12}
\caption{
Comparison with representative agent benchmarks.
\cmark indicates full support, \pmark indicates partial support or a different formulation, and \xmark indicates unsupported or not applicable.
``Avg. Tools / Env.'' is normalized by the number of environments, applications, domains, or servers when such grouping is available.
``Var.'' denotes task-dependent values, ``N/A'' denotes benchmarks without tool-based environments, and ``N/R'' denotes values not reliably reported in public sources.
}
\label{tab:related_work_comparison}

\begin{adjustbox}{max width=\textwidth}
\begin{tabular}{lccccccccc}
\toprule
\multirow{2}{*}{\raisebox{-0.8ex}{\textbf{Benchmark}}}
& \multicolumn{3}{c}{\textbf{Environment \& Task}}
& \multicolumn{2}{c}{\textbf{Domain \& Capability}}
& \multicolumn{4}{c}{\textbf{Interaction}} \\
\cmidrule(lr){2-4}
\cmidrule(lr){5-6}
\cmidrule(lr){7-10}
& \textbf{Tasks}
& \makecell{\textbf{Avg. Tools}\\\textbf{/ Env.}}
& \makecell{\textbf{State}\\\textbf{Init.}}
& \makecell{\textbf{Domains}\\\textbf{Taxonomy}}
& \makecell{\textbf{Capability}\\\textbf{Taxonomy}}
& \makecell{\textbf{Approx}\\\textbf{Turns}}
& \textbf{Persona}
& \makecell{\textbf{Interaction}\\\textbf{Mode}}
& \textbf{Workspace} \\
\midrule

Toolathlon~\cite{li2025tool}
& 108
& 18.9
& \cmark
& \pmark~(32 apps)
& \pmark
& 26.8
& \xmark
& Multi
& \cmark \\

VitaBench~\cite{he2025vitabench}
& 400
& 66
& \cmark
& \cmark~(3)
& \pmark
& 50--100
& \cmark
& Multi
& \pmark \\

GAIA2~\cite{froger2026gaia2}
& 800 / 1120
& 101
& \cmark
& \cmark~(10)
& \cmark~(7)
& 22.5
& \pmark
& Multi
& \pmark \\

ToolSandbox~\cite{lu2025toolsandbox}
& 1032
& 34.0
& \cmark
& \cmark~(11)
& \pmark
& 13.9
& \pmark
& Both
& \xmark \\

ACEBench~\cite{chen2025acebench}
& 2000
& 66.7
& \pmark
& \cmark~(8/68)
& \pmark
& 1.7
& \pmark
& Both
& \xmark \\

$\tau^2$-Bench~\cite{barres2025tau}
& 279
& 12.7
& \cmark
& \cmark~(3)
& \pmark
& 30--80
& \pmark
& Multi
& \xmark \\

$\tau$-knowledge~\cite{shi2026tauknowledgeevaluatingconversationalagents}
& 97
& 51
& \cmark
& \pmark~(1)
& \pmark
& N/R
& \cmark
& Both
& \pmark \\

BFCL v3~\cite{patil2024gorilla}
& 4441
& Var.
& \cmark
& \pmark
& \pmark
& Var.
& \pmark
& Both
& \pmark \\

BFCL v4~\cite{patil2025berkeley}
& 5088
& Var.
& \cmark
& \pmark
& \pmark
& Var.
& \pmark
& Both
& \pmark \\

MCP-Atlas~\cite{bandi2026mcp}
& 1000
& 6.1
& \pmark
& \pmark~(36 servers)
& \pmark
& 3--6 calls
& \xmark
& Single
& \pmark \\

DeepPlanning~\cite{zhang2026deepplanning}
& 360
& 12.0
& \cmark
& \cmark~(2)
& \cmark~(3)
& Var.
& \xmark
& Single
& \xmark \\

GDPval~\cite{patwardhan2025gdpval}
& 1320
& N/A
& \xmark
& \cmark~(44/9)
& \xmark
& 1
& \xmark
& Single
& \pmark \\

Vending-Bench~\cite{backlund2025vending}
& 1
& 10
& \cmark
& \xmark
& \xmark
& 3k--6k msg.
& \pmark
& Multi
& \xmark \\

Claw-Eval~\cite{ye2026claw}
& 300
& Var.
& \cmark
& \cmark~(9)
& \cmark~(3)
& Var.
& \pmark
& Both
& \cmark \\

ClawMark~\cite{meng2026clawmark}
& 100
& 5.0
& \cmark
& \cmark~(13)
& \pmark
& 1--3 days
& \pmark
& Multi
& \cmark \\

QwenClawBench~\cite{qwenclawbench1.1}
& 100
& N/R
& \cmark
& \cmark~(8)
& \xmark
& N/R
& \xmark
& Single
& \cmark \\

\midrule
\textbf{OmniaBench}
& \textbf{644 / 1431}
& \textbf{65.32}
& \cmark
& \textbf{\cmark~(90/354)}
& \textbf{\cmark~(10)}
& ~100
& \cmark
& Both
& \cmark \\

\bottomrule
\end{tabular}
\end{adjustbox}
\end{table*}

Recent evaluation of LLM-based agents has gradually shifted from static question answering toward tool use, environment interaction, and task execution. A line of foundational tool-use benchmarks studies whether models can select appropriate tools, generate valid arguments, compose multiple function calls, and satisfy execution constraints. Representative works such as BFCL~\cite{patil2025berkeley}, ACEBench~\cite{chen2025acebench}, and ToolSandbox~\cite{lu2025toolsandbox} have advanced standardized evaluation from the perspectives of function calling, compositional tool use, stateful execution, and different invocation patterns. These benchmarks provide important foundations for evaluating agentic tool use, enabling systematic analysis of model capabilities in tool selection, argument construction, and execution control.

As agent tasks move closer to real-world applications, recent benchmarks further introduce richer tool ecosystems, interactive environments, dynamic states, and more diverse task forms. MCP-Atlas~\cite{bandi2026mcp} and Toolathlon~\cite{li2025tool} extend agent evaluation to realistic tool collections and application ecosystems, emphasizing task completion across tools, services, or applications. $\tau^2$-Bench~\cite{barres2025tau} and $\tau^3$-Bench~\cite{shi2026tauknowledgeevaluatingconversationalagents} incorporate simulated users, domain rules, and stateful databases, allowing agents to complete tasks through multi-turn interactions in service-oriented scenarios. VitaBench~\cite{he2025vitabench}, GAIA2~\cite{froger2026gaia2}, and DeepPlanning~\cite{zhang2026deepplanning} further examine agent performance in real-life services, open-ended tasks, long-horizon planning, and capability-compositional settings. GDPval~\cite{patwardhan2025gdpval} and Vending-Bench~\cite{backlund2025vending} explore model capabilities in professional tasks and long-running business decision-making, respectively. In parallel, Claw-Eval~\cite{ye2026claw}, ClawMark~\cite{meng2026clawmark}, and QwenClawBench~\cite{qwenclawbench1.1} focus on workspace or computer-use agents, highlighting task execution over file systems, documents, spreadsheets, code, and multi-service sandboxes. Together, these benchmarks have substantially expanded the scope of general agent evaluation from different perspectives.

Building on these efforts, OmniaBench aims to provide a general-agent benchmark with broader domain coverage, multiple interaction modes, and multi-dimensional capability analysis. Single-turn tasks are useful for evaluating whether a model can understand a user request, plan an execution strategy, and complete the goal in one attempt, while multi-turn tasks emphasize clarification, feedback utilization, state maintenance, and iterative correction during interaction. Both forms are important for comprehensive general-agent evaluation. As summarized in Table~\ref{tab:related_work_comparison}, OmniaBench constructs a large-scale domain taxonomy from real-world application resources and integrates capability taxonomy, stateful environments, persona design, and diverse interaction modes. By covering single-turn, multi-turn, workspace, coding, and tool-use tasks within a unified benchmark, OmniaBench provides a complementary perspective for analyzing the capability boundaries of current general agents across diverse scenarios.

\section{OmniaBench: Evaluating General Agents in Diverse Scenarios}

\subsection{Formulation}
\label{sec:formulation}

A general-agent task can be naturally formulated as a partially observable Markov decision process (POMDP), where an agent interacts with an environment through observations and actions while only having access to partial information about the underlying state. Formally, we denote the interactive process as
\[
\mathcal{P} = (\mathcal{S}, \mathcal{O}, \mathcal{A}, \mathcal{T}, \Omega, \rho_0),
\]
where \(\mathcal{S}\) is the latent state space, \(\mathcal{O}\) is the observation space, \(\mathcal{A}\) is the action space, \(\mathcal{T}: \mathcal{S} \times \mathcal{A} \rightarrow \Delta(\mathcal{S})\) is the transition function, \(\Omega: \mathcal{S} \times \mathcal{A} \rightarrow \Delta(\mathcal{O})\) is the observation function, and \(\rho_0\) is the initial state distribution. A latent state \(s_t \in \mathcal{S}\) specifies the complete task configuration at step \(t\), including user-side information, environment records, tool states, workspace contents, and intermediate execution results. The agent, however, only receives an observation \(o_t \in \mathcal{O}\), such as a user message, a tool response, a retrieved document, a file state, or an environment feedback signal.

At each step \(t\), the agent conditions its decision on the interaction history
\[
h_t = (o_0, a_0, o_1, a_1, \ldots, o_t),
\]
rather than the latent state \(s_t\). Given \(h_t\), an agent policy \(\pi_\theta\) produces an action
\[
a_t \sim \pi_\theta(\cdot \mid h_t).
\]
The action space of a general agent is heterogeneous. It may include natural-language responses, clarification questions, structured tool calls, retrieval actions, code execution, file operations, and task termination signals. In OmniaBench, workspace-related operations such as file manipulation or code execution are exposed through controlled tool interfaces, so that their results are returned as tool observations while maintaining a unified and safe execution protocol. We therefore write
\[
\mathcal{A}
=
\mathcal{A}_{\mathrm{msg}}
\cup
\mathcal{A}_{\mathrm{tool}}
\cup
\mathcal{A}_{\mathrm{final}},
\]
where \(\mathcal{A}_{\mathrm{msg}}\) denotes user-facing messages, \(\mathcal{A}_{\mathrm{tool}}\) denotes external tool invocations including workspace and code tools, and \(\mathcal{A}_{\mathrm{final}}\) denotes task finalization actions used in single-turn settings.

\begin{figure*}[t]
\vspace{-2em}
    \centering
    \includegraphics[width=1\linewidth]{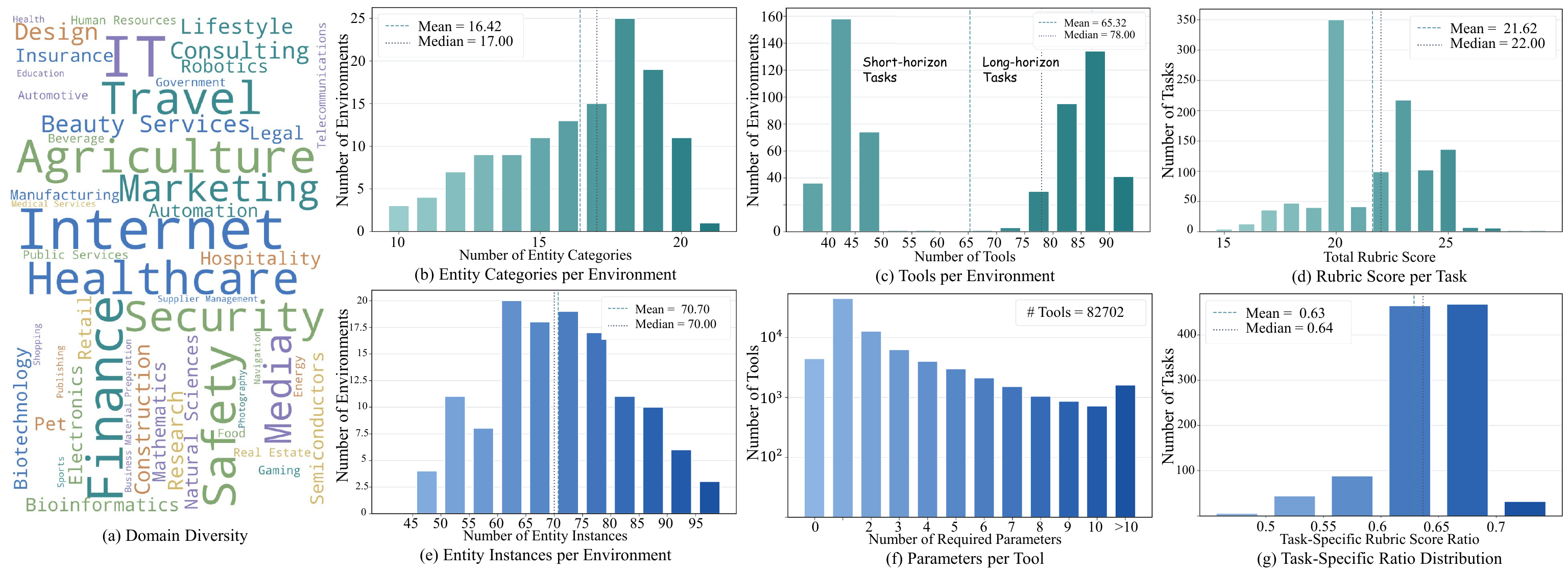}
    \caption{\textbf{Statistics of OmniaBench.}
Comprehensive statistics of the benchmark, including domain diversity, environment complexity, tool availability, task difficulty, and rubric characteristics.}
    \label{fig:env_statics}
\end{figure*}

Once \(a_t\) is executed, the environment evolves according to
\[
s_{t+1} \sim \mathcal{T}(\cdot \mid s_t, a_t),
\qquad
o_{t+1} \sim \Omega(\cdot \mid s_{t+1}, a_t).
\]
This process captures the core difficulty of agentic tasks: the agent must infer hidden task-relevant information from partial observations, choose actions under uncertainty, and update its plan as new evidence is revealed.

User interaction can be incorporated into the same POMDP view by treating the user as part of the environment dynamics. In multi-turn settings, the next observation may include user feedback generated according to the task goal, user constraints, the current state, and the previous dialogue. Therefore, the agent must not only execute actions, but also clarify underspecified requests, track evolving constraints, and revise its strategy during interaction. In this setting, task completion is determined through the subsequent user response or a user-side stop signal rather than an explicit finalization action from the agent. Single-turn tasks can be viewed as a special case in which the main task information is provided in the initial observation, and the agent completes the goal within one response or a limited sequence of tool calls, typically ending with a final submission action.

A complete execution induces a trajectory
\[
\Gamma =
(o_0, a_0, o_1, a_1, \ldots, o_T, a_T),
\]
where \(T\) denotes the termination step. For single-turn tasks, termination is typically triggered by a final submission action \(a_T \in \mathcal{A}_{\mathrm{final}}\). For multi-turn tasks, termination is triggered by the user-side stop signal after the interaction reaches a satisfactory state. In both settings, execution may also stop when the task reaches the maximum turn limit, exceeds the tool-call budget, or triggers an environment-level failure condition.

Given a task goal \(g\) and an evaluation protocol \(\mathcal{V}\), the trajectory is assessed by
\[
\mathrm{Eval}_{\mathcal{V}}(\Gamma, g) \rightarrow (y, r),
\]
where \(y \in \{0,1\}\) is the binary success label and \(r\) denotes rubric scores that capture fine-grained task completion criteria. Task completion is generally assessed through rubric-based judgment, while programmatically verifiable tasks are equipped with \VerifyCode{} to determine the pass/fail signal based on the trajectory and the final observation. Under this formulation, general-agent evaluation measures not only the final task outcome, but also the agent's ability to make effective sequential decisions under partial observability.

\subsection{Taxonomy}
\label{sec:taxonomy}

OmniaBench is organized by a three-view taxonomy that connects
real-world application coverage with agent capability analysis. The
domain taxonomy captures where a task arises, covering consumer-facing, business-facing, and employee-oriented settings. The capability taxonomy characterizes the general-agent abilities required to solve a task, while the atomic difficulty taxonomy describes compositional sources of task complexity across the user, environment, tool, and interaction sides. Table~\ref{tab:taxonomy_overview} summarizes the sources and scales of the three taxonomy views. Detailed definitions for the ten capability dimensions and the eight atomic difficulty factors are provided in Appendices~\ref{appendix:capability-taxonomy} and \ref{appendix:atomic-difficulty}, respectively.

\begin{table}[t]
\centering
\small
\setlength{\tabcolsep}{4pt}
\renewcommand{\arraystretch}{1.10}
\caption{Overview of the taxonomy structure.}
\label{tab:taxonomy_overview}
\begin{tabularx}{\linewidth}{@{}l l X c@{}}
\toprule
\textbf{View} & \textbf{Type} & \textbf{Source / Basis} & \textbf{Scale} \\
\midrule
\multirow{4}{*}{Domain}
& ToC & Multiple app stores and consumer-facing app categories & 22 L1 / 101 L2 \\
& ToB & GDPval-style~\cite{patwardhan2025gdpval} tasks and industry classifications & 38 L1 / 186 L2 \\
& ToE & Industry-general employee tasks from GDPval and industry templates & 30 L1 / 67 L2 \\
\cmidrule(lr){2-4}
& Total & Normalized real-world scenario taxonomy & 90 L1 / 354 L2 \\
\midrule
Capability
& Dims. & General-agent execution requirements & 10 dims. \\
\midrule
Atomic Difficulty
& Factors & Eight challenge factors across user, environment, tool use, and interaction & 8 factors \\
\bottomrule
\end{tabularx}
\end{table}

\subsection{Data Construction}
\label{sec:data_construction}

\begin{figure*}[t]
    \centering
    \includegraphics[width=\textwidth]{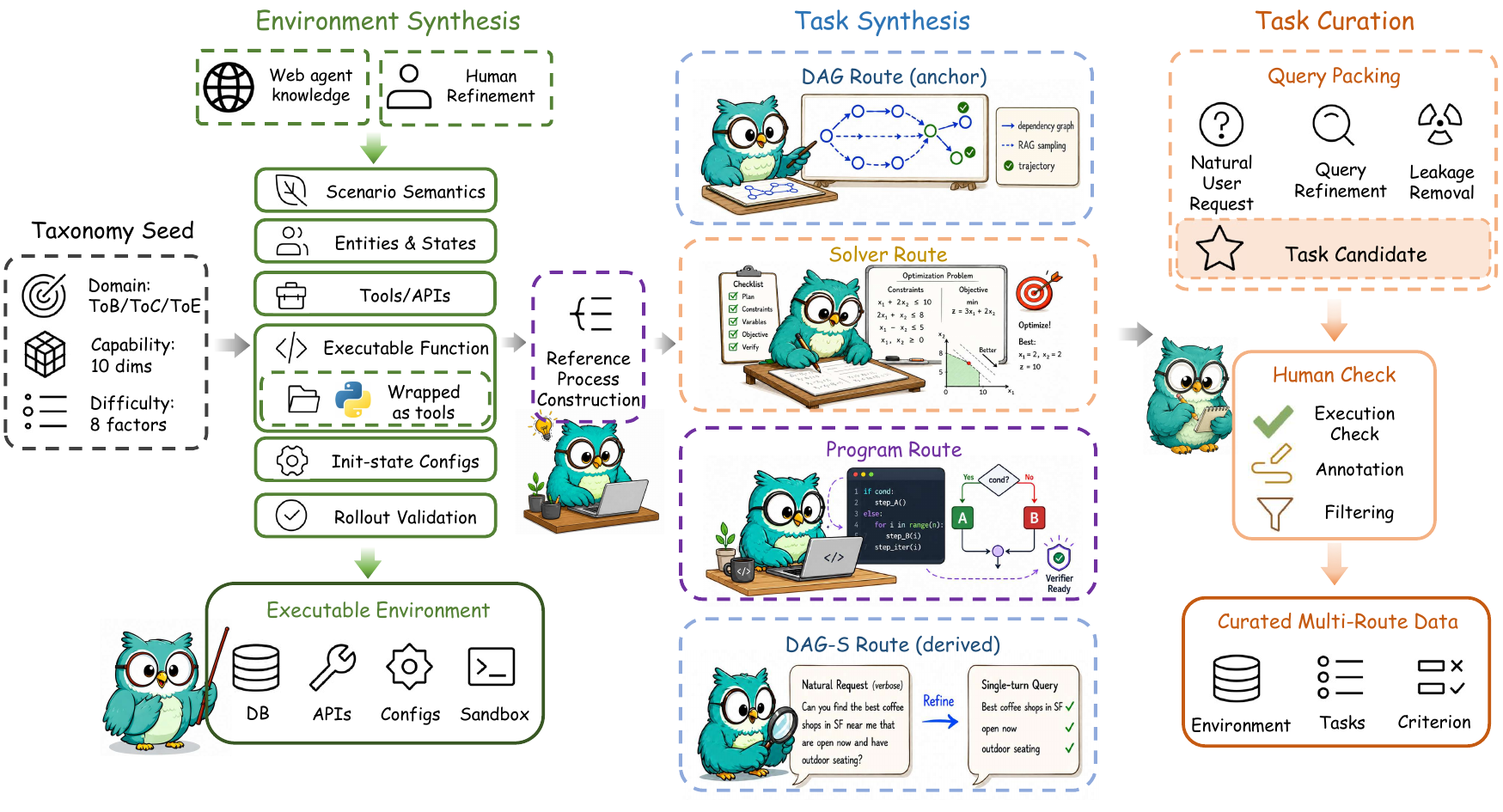}
    \caption{\textbf{Overview of the data construction pipeline.}
    Starting from the taxonomy, domain knowledge collected by a Web agent is incorporated into environment and task construction to produce four complementary forms of data, which are subsequently curated through human annotation to obtain the final benchmark data.}
    \label{fig:data_construction_pipeline}
    \vspace{-1em}
\end{figure*}

We first construct executable environments from the taxonomy-grounded scenario space as shown in Figure~\ref{fig:data_construction_pipeline}. For each level-2 domain category, we use a Web agent to collect domain-specific information from real applications, product functions, business workflows, and user needs, and further calibrate the collected knowledge through human refinement. The resulting external knowledge is injected into environment construction, where we infer the entities, attributes, state variables, and domain constraints required to support the target domain. These specifications are then translated into instantiable Python environment classes. For each environment, we generate domain-specific operations, including information-query tools and state-changing tools, together with executable function implementations, tool schemas, and initialization configurations. For file-system and coding tasks, we additionally provide controlled sandbox environments, where predefined file-operation tools and a Python executor are wrapped as adapters and exposed through the same tool interface as ordinary environment tools. We then apply syntax checking, tool-call rollout, and state-consistency validation to filter out non-executable or unstable environments, ensuring each environment supports realistic state reading, state mutation, tool interaction, and controlled workspace execution.

Given the constructed environments, we synthesize tasks with domain knowledge injected throughout the generation process. The Web-agent-collected knowledge is used to guide task construction so that generated tasks reflect natural user goals, domain constraints, and operational workflows in the corresponding domain. For the DAG-based route, we build a tool dependency graph from the environment tool set, sample executable tool chains or DAGs according to dependency relations, and then generate task requests, reference trajectories, and final states conditioned on environment states. The generated tasks are further validated through real environment execution, parameter repair, and task--trajectory--state consistency checking. To evaluate information retrieval, aggregation, and comparison abilities, we also expand instantiated information in the initial states, introducing richer candidate objects, distractors, cross entity relations, and aggregatable histories. Based on this backbone, we construct four complementary data routes: \textit{DAG} for multi-turn stateful interaction and tool-chain execution; \textit{DAG-S} for single-turn tasks obtained through query refinement over DAG-based tasks; \textit{Solver-based} tasks for selection, scheduling, allocation, and optimization scenarios using implicit solver-guided or explicit solver-anchored synthesis; and \textit{Program-based} tasks for complex procedural reasoning with branching, iteration, task-program synthesis, execution debugging, and query refinement.

Finally, we construct evaluation signals for different data types.
The task of DAG, Solver, DAG-S is equipped with rubric criteria covering both general
completion quality and task-specific requirements, followed by
consistency checks on rubric format, score allocation, and validity. For the Program-based route, we construct \VerifyCode{} to determine binary success based on trajectories and final observations. The generated environments, tasks, trajectories, rubrics, and verifiers are further refined through human annotation and curation, focusing on solvability, semantic consistency, evaluation validity, and ambiguity reduction.

\noindent\textbf{Challenging set construction.}
To reduce evaluation cost and mitigate potential contamination of the full set after public release while preserving broad scenario coverage, we derive a challenging set from the full collection. For the DAG route, we select the task with the longest reference tool chain generated during data synthesis for every level-2 domain, yielding 354 DAG tasks in total. For the remaining three routes, we retain most of their validated instances after excluding a limited number of low-complexity cases, resulting in 200 DAG-S tasks, 60 Solver tasks, and 30 Program tasks. The final challenging set contains \textbf{644 tasks} in total. Statistics for the challenging set are reported in Table~\ref{tab:multi_category_overall_statistics}, while statistics for the full \textbf{1,431 tasks} are provided in Appendix Table~\ref{tab:multi_category_overall_statistics_appendix}. The resulting benchmark supports cost-efficient leaderboard evaluation, route-wise comparison, capability diagnosis, and error analysis.

\subsection{Evaluation}

\label{sec:evaluation_protocol}
OmniaBench adopts a unified trajectory-based evaluation framework with route-specific verification signals. Except for Program-based data, each task is assessed by multiple rubric items covering distinct objectives and constraints, which are aggregated into task-level scores and pass/fail outcomes. Program-based tasks are evaluated with \VerifyCode{}, which directly determines the binary outcome from the complete trajectory and final observation. For multi-turn tasks, we employ persona-grounded user simulators (detailed in Appendix~\ref{app:persona-design}) and control interaction difficulty through user preferences, communication styles, information-disclosure strategies, and varying levels of conversational complexity. Filesystem and coding tasks are executed in controlled sandboxes, where predefined file tools and a Python executor are exposed through the same interface as ordinary tools to ensure safety and reproducibility. After each run, the system generates a visual HTML report that consolidates the dialogue, tool calls, environment feedback, final observation, rubric scores, and failure reasons, facilitating manual inspection and case-level analysis.

\begin{table*}[htbp]
\centering
\caption{\textbf{Statistics of the challenging set across task categories.}
All values are averaged over tasks within each category except \#Tasks. 
Note that the Program route uses a binary \VerifyCode{} verifier instead of multi-item textual rubrics.
Statistics for the full set are reported in Appendix Table~\ref{tab:multi_category_overall_statistics_appendix}.}

\label{tab:multi_category_overall_statistics}
\setlength{\tabcolsep}{5pt}
\renewcommand{\arraystretch}{1.05}
\small

\begin{adjustbox}{max width=\textwidth}
\begin{tabular}{lrrrrr}
\toprule
\rowcolor{HeaderBlue}
\textbf{Metric}
& \textbf{DAG}
& \textbf{DAG-S}
& \textbf{Solver}
& \textbf{Program}
& \textbf{Overall} \\
\midrule

\rowcolor{GroupGray}
\multicolumn{6}{l}{\textbf{Dataset Size}} \\
\#Tasks
& 354
& 200
& 60
& 30
& \textbf{644} \\

\addlinespace[0.25em]
\rowcolor{GroupGray}
\multicolumn{6}{l}{\textbf{Task}} \\
Task length
& 953.2
& 2922.0
& 1450.5
& 4443.6
& \textbf{1773.6} \\

\addlinespace[0.25em]
\rowcolor{GroupGray}
\multicolumn{6}{l}{\textbf{Environment}} \\
\#Tools
& 65.3
& 56.9
& 8.2
& 52.2
& \textbf{59.0} \\

\#Entity tables
& 16.4
& 13.1
& 6.7
& 12.6
& \textbf{13.7} \\

\#Entities
& 70.7
& 205.7
& 111.1
& 312.4
& \textbf{125.2} \\


\addlinespace[0.25em]
\rowcolor{GroupGray}
\multicolumn{6}{l}{\textbf{Rubric or VerifyCode}} \\

\# Items
& 9.0
& 13.0
& 5.8
& 1
& \textbf{10.0} \\



Item points
& 21.6
& 32.4
& 5.8
& 1
& \textbf{23.6} \\

Description length
& 133.9
& 251.8
& 429.3
& \VerifyCode{}
& \textbf{201.2} \\

\bottomrule
\end{tabular}
\end{adjustbox}

\end{table*}

\section{Experiments}

\definecolor{lightgray}{gray}{0.93}
\definecolor{lightblue}{RGB}{220,235,250}

\begin{table*}[t]
\centering
\small

\setlength{\tabcolsep}{3.5pt}
\renewcommand{\arraystretch}{1.15}

\begin{tabular}{lcccccccccc}
\toprule

\multirow{2}{*}{\textbf{Models}}
&
\multicolumn{2}{c}{\textbf{DAG}}
&
\multirow{2}{*}{\textbf{Solver}}
&
\multirow{2}{*}{\textbf{Program}}
&
\multirow{2}{*}{\textbf{DAG-S}}
&
\multirow{2}{*}{\textbf{ToB}}
&
\multirow{2}{*}{\textbf{ToC}}
&
\multirow{2}{*}{\textbf{ToE}}
&
\multirow{2}{*}{\textbf{ToolSteps}}
&
\multirow{2}{*}{\textbf{Overall}}
\\

&
\textbf{Pass@1}
&
\textbf{UserTurns}
&
&
&
&
&
&
&
&
\\

\midrule

\rowcolor{lightgray}
\multicolumn{11}{c}{
\textit{\textbf{Closed-source Models}}
}
\\
\midrule

Claude-Sonnet-5
&\textbf{57.34}&9.97
&56.67&63.33&60.50
&\textbf{60.68}&\textbf{60.47}&48.11
&64.16
&\cellcolor{lightblue}\textbf{58.54}
\\

GPT-5.6-sol
&55.37&7.52
&\textbf{65.00}&50.00&59.00
&57.59&59.07&51.89
&38.73
&\cellcolor{lightblue}57.14
\\

GPT-5.5
&54.80&7.27
&38.33&60.00&64.50
&58.20&53.02&\textbf{58.49}
&52.45
&\cellcolor{lightblue}56.52
\\

Claude-Opus-4.7
&53.39&7.60
&43.33&63.33&57.50
&56.66&51.63&51.89
&44.11
&\cellcolor{lightblue}54.19
\\

Qwen3.7-Max
&48.59&6.49
&51.67&\textbf{66.67}&48.50
&48.30&48.84&55.66
&47.96
&\cellcolor{lightblue}49.69
\\

Gemini-3.5-Flash
&54.24&6.16
&36.67&30.00&35.50
&46.75&46.98&39.62
&61.82
&\cellcolor{lightblue}45.65
\\

Doubao-Seed-2.1-Pro
&54.80&6.85
&25.00&36.67&27.50
&43.03&40.93&45.28
&44.83
&\cellcolor{lightblue}42.70
\\

Gemini-3.1-Pro
&50.28&7.33
&35.00&30.00&29.00
&42.11&39.53&42.45
&49.19
&\cellcolor{lightblue}41.30
\\

Doubao-Seed-2.0-Pro
&33.90&8.14
&21.67&36.67&5.00
&22.60&25.12&25.47
&29.76
&\cellcolor{lightblue}23.91
\\

GPT-5.4-mini
&29.10&8.84
&0.00&6.67&3.00
&15.48&18.14&20.75
&37.12
&\cellcolor{lightblue}17.24
\\

\midrule

\rowcolor{lightgray}
\multicolumn{11}{c}{
\textit{\textbf{Open-source Models}}
}
\\
\midrule

GLM-5.2
&54.80&6.83
&26.67&60.00&\textbf{69.00}
&57.28&55.81&57.55
&57.08
&\cellcolor{lightblue}56.83
\\

DeepSeek-V4-Pro
&52.54&6.23
&36.67&53.33&63.50
&54.18&54.88&54.72
&58.08
&\cellcolor{lightblue}54.50
\\

Kimi-K2.6
&49.72&7.09
&45.00&63.33&57.50
&54.80&49.30&50.94
&50.70
&\cellcolor{lightblue}52.33
\\

DeepSeek-V4-Flash
&51.13&6.18
&28.33&43.33&51.00
&45.20&51.63&52.83
&59.96
&\cellcolor{lightblue}48.60
\\

GLM-5
&50.85&7.50
&10.00&53.33&38.50
&47.68&37.67&41.51
&57.59
&\cellcolor{lightblue}43.32
\\

Qwen3.6-27B
&48.02&7.66
&28.33&46.67&35.00
&43.65&38.60&44.34
&46.62
&\cellcolor{lightblue}42.08
\\

Qwen3.5-397B-A17B
&50.56&7.42
&31.67&50.00&26.00
&44.27&35.35&43.40
&54.67
&\cellcolor{lightblue}41.15
\\

Qwen3.6-35B-A3B
&43.22&7.46
&20.00&36.67&32.00
&41.80&31.63&34.91
&53.28
&\cellcolor{lightblue}37.27
\\

Qwen3.5-35B-A3B
&38.70&6.31
&25.00&30.00&9.50
&29.41&21.40&36.79
&60.66
&\cellcolor{lightblue}27.95
\\

Qwen3.5-9B
&41.53&7.06
&16.67&23.33&8.00
&30.96&20.93&33.02
&66.19
&\cellcolor{lightblue}27.95
\\

Qwen3-235B-A22B
&35.31&7.16
&0.00&6.67&1.00
&21.36&20.00&16.04
&43.21
&\cellcolor{lightblue}20.03
\\

Qwen3-30B-A3B
&16.67&9.13
&0.00&0.00&0.00
&9.91&6.51&12.26
&29.63
&\cellcolor{lightblue}9.16
\\

\bottomrule

\end{tabular}

\caption{\textbf{Benchmark results across different evaluation routes.}
Pass@1 performance across four task construction routes: DAG, Solver, Program, and DAG-S. 
Overall denotes the total Pass@1 scores over all  tasks of four routes. The reasoning effort and other parameters of models are illustrated in Subsection~\ref{subsec:effort}. }

\label{tab:benchmark}

\end{table*}

\subsection{Experimental Setups}
\label{sec:experimental_setups}

\textbf{Models.}
We evaluate a collection of closed and open source models on OmniaBench. The closed source models include OpenAI's GPT-5.6-Sol, GPT-5.5 and GPT-5.4-Mini, Anthropic's Claude-Sonnet-5, Claude-Opus-4.7, Google's Gemini-3.5-Flash and Gemini-3.1-Pro-Preview, Qwen3.7-Max, Doubao-Seed-2.1-Pro-2026-06-28, and Doubao-Seed-2.0-Pro-2026-02-15. The open-source models include Kimi-K2.6, GLM-5.2, DeepSeek-V4-Pro, DeepSeek-V4-Flash, Qwen3.5-397B-A17B, Qwen3.6-27B, GLM-5, Qwen3.6-35B-A3B, Qwen3.5-35B-A3B, Qwen3.5-9B, Qwen3-235B-A22B-THINKING-2507, and Qwen3-30B-A3B-THINKING-2507\footnote{All evaluations were conducted from June 29 to July 15, 2026, using the provider APIs available during this period.}.

\textbf{Metrics.}
We use Pass@1 as the primary leaderboard metric and separately report task success rates on the DAG, Solver, Program, and DAG-S subsets. For rubric-based tasks, each task is associated with multiple rubric items weighted from 1 to 3 points. Each item receives either its full score or zero, and a task is considered successful only when all rubric items are satisfied. Program tasks are instead assigned binary outcomes directly by \VerifyCode{}. The \textbf{Overall} score is computed as the Pass@1 over all four routes, obtained by dividing the total number of successfully completed tasks by the total number of tasks in the challenging set (\(644\)). The ToB, ToC, and ToE scores are computed by dividing the number of successfully completed tasks in each split by the total number of tasks in that split, with domain-level scores calculated analogously. \textbf{Capability} scores are computed as the average Pass@1 over tasks involving the corresponding capability. We additionally report \textbf{UserTurns}, the average number of user turns in DAG multi-turn tasks, including the initial user request and both successful and unsuccessful trajectories, and \textbf{ToolSteps}, the average number of tool-call steps across all trajectories, including failed or prematurely terminated runs. We quantify user model robustness using Kendall's \(\tau\) over model rankings and the average coefficient of variation (CV) of model scores across user simulators. Rubric-judge stability is evaluated separately using agreement-based statistics, as detailed in Appendix~\ref{appendix:judge-exp}. Separate experiments further report Pass@8 and Pass\textsuperscript{\^{}}8, measuring whether at least one or all eight independent runs succeed~\cite{yao2024tau}, respectively.

\textbf{Implementation Details.}
\label{subsec:effort}
We use Qwen3.5-397B-A17B for environment and task synthesis during data construction. For multi-turn evaluation, DeepSeek-V4-Pro with thinking disabled serves as both the user simulator and the rubric judge. The user simulator terminates an interaction by emitting the special signal \texttt{\#\#\#STOP\#\#\#}, whereas single-turn tasks terminate when the evaluated model submits its final response. Temperature is set to 0 whenever supported; for the Claude and Gemini series, for which the official guidance recommends retaining the default setting, we use the default temperature of 1.0. GPT models use \texttt{high} reasoning effort, Claude series use the \texttt{max} reasoning budget, other models use their max effort mode if exist, and explicit thinking is enabled for all other models that support it. Each trajectory permits at most 200 tool-call steps, rather than 200 model-output turns. We allow up to 20 cumulative retries for unexpected environment-side tool exceptions; API-level failures are handled separately and do not count toward this limit. All tools follow the OpenAI function-calling schema, with consistent tool definitions across models.

\subsection{Main Results}

\begin{figure}[t]
    \centering
    \vspace{-2em}
    \begin{subfigure}[t]{0.40\linewidth}
        \centering
        \vspace{0pt}
        \includegraphics[width=\linewidth]{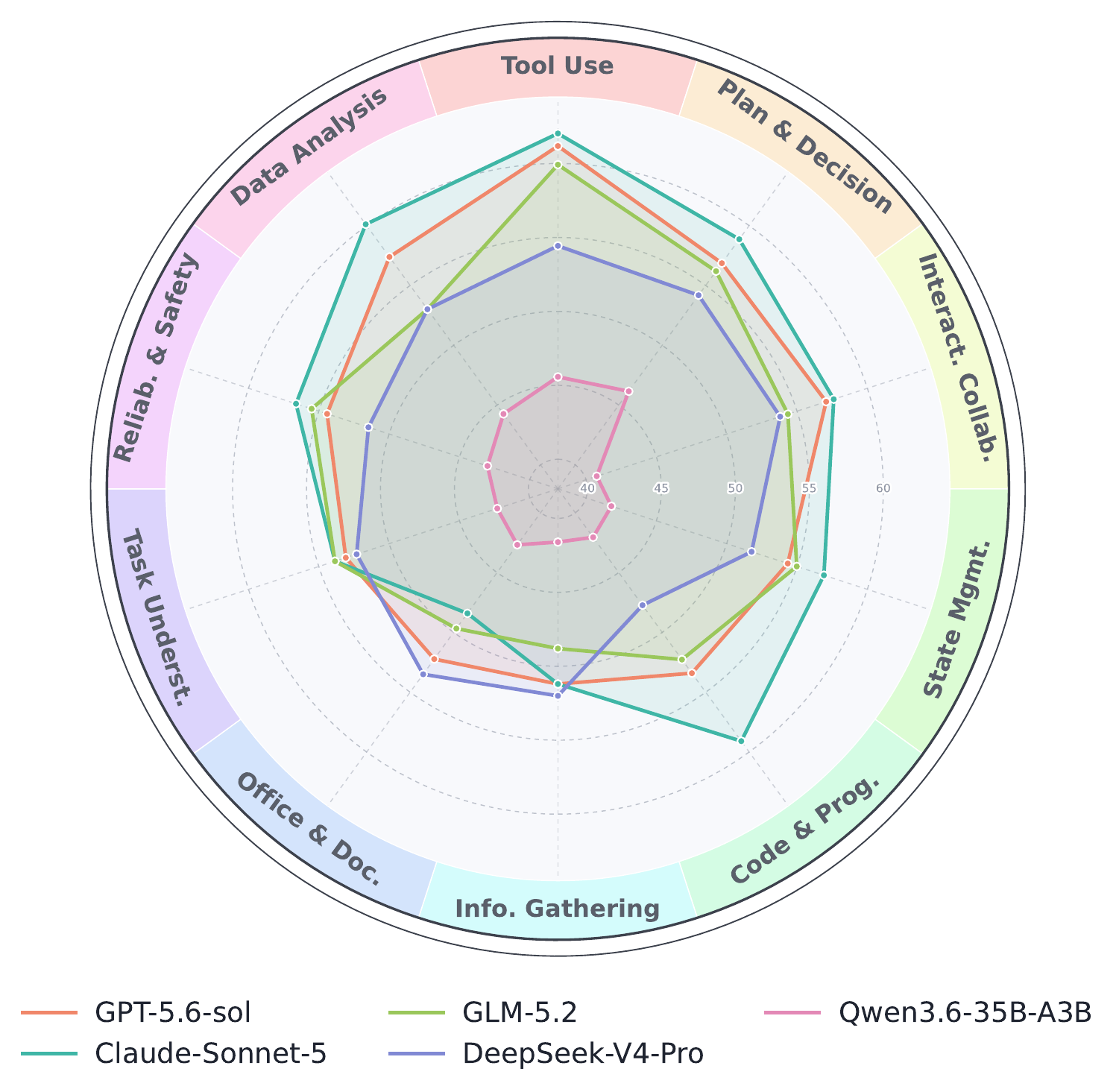}
        \caption{Capability profiles.}
        \label{fig:capability_radar}
    \end{subfigure}
    \hfill
    \begin{subfigure}[t]{0.58\linewidth}
        \centering
        \vspace{0pt}
        \includegraphics[width=\linewidth]{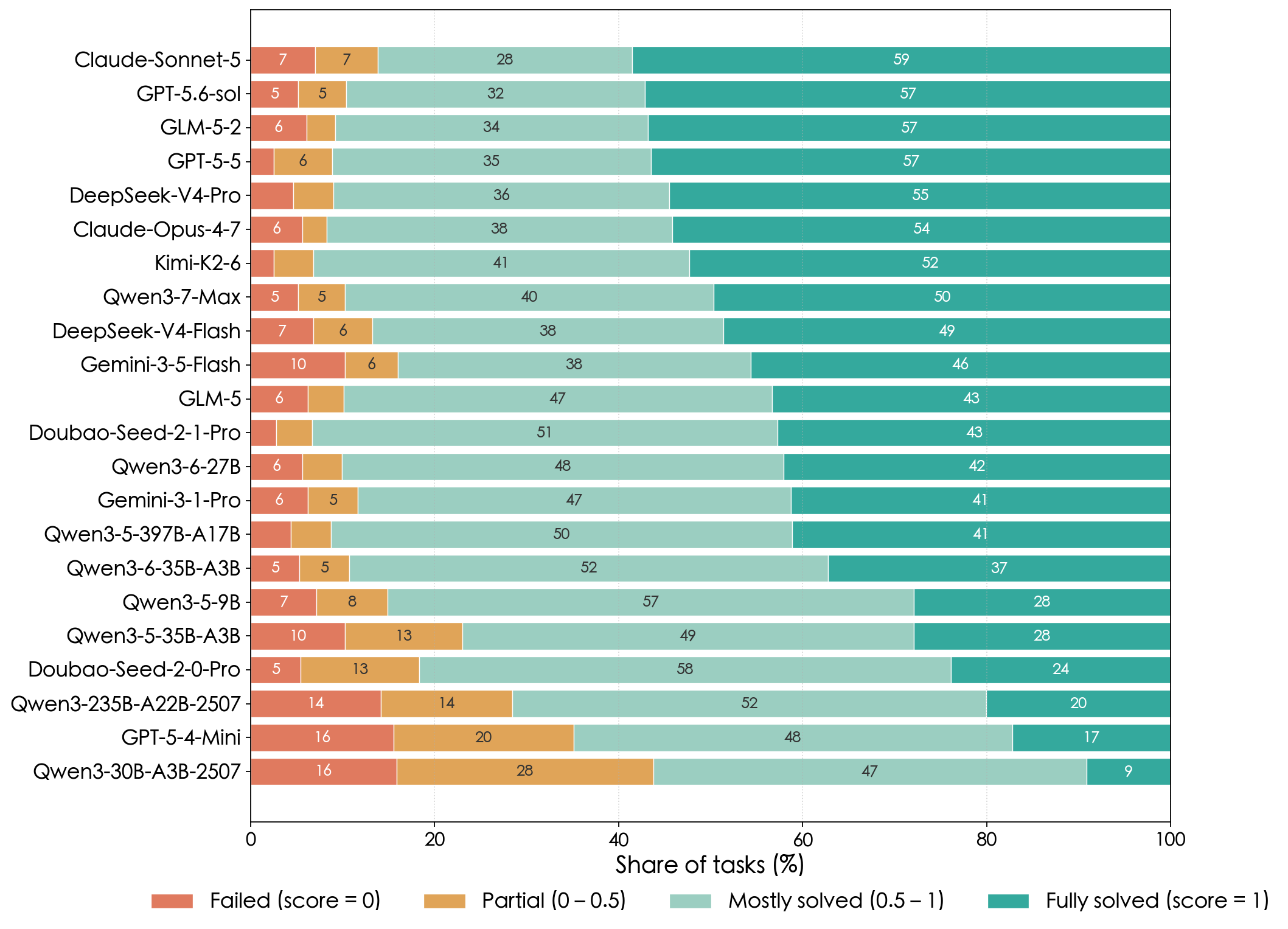}
        \caption{Per-task score distribution.}
        \label{fig:score_distribution}
    \end{subfigure}

\caption{
\textbf{Capability and score analyses.}
(a) Capability profiles of representative models across the ten capability dimensions. For each model and capability, the reported value is computed as the arithmetic mean of the model's normalized task scores over all tasks annotated with the corresponding capability label. Details for capability dimensions are provided in Appendix~\ref{appendix:capability-taxonomy}.
(b) Distribution of normalized per-task scores for all models on the
leaderboard.
}
    \label{fig:capability_score_analysis}
\end{figure}

\begin{figure}[t]
    \centering

    \begin{subfigure}[b]{0.40\linewidth}
        \centering
        \raisebox{1.2em}{
        \includegraphics[width=\linewidth]{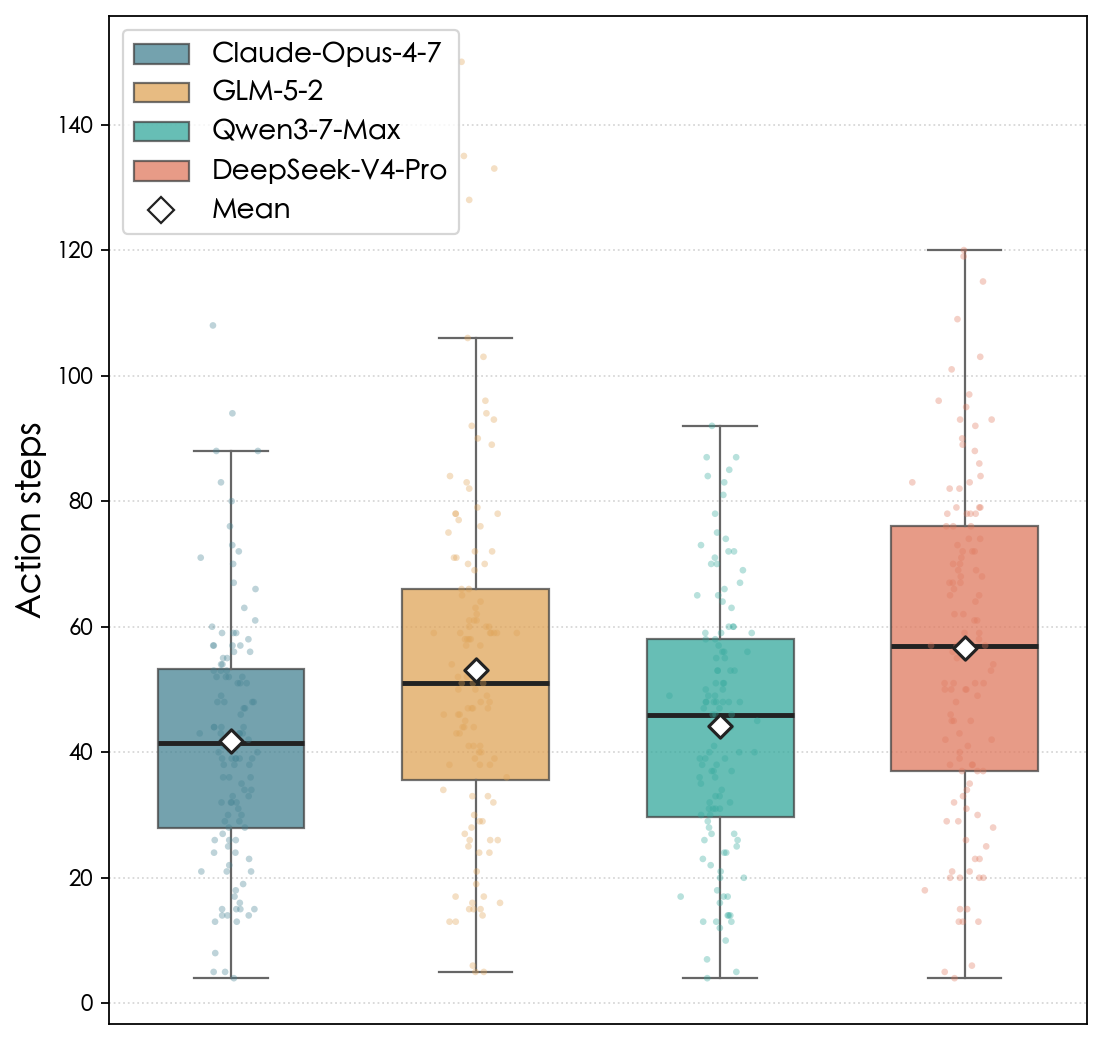}
        }
        \caption{Step efficiency on the commonly-solved set.}
        \label{fig:step_efficiency}
    \end{subfigure}
    \hspace{0.1\linewidth}
    \begin{subfigure}[b]{0.43\linewidth}
        \centering
        \includegraphics[width=\linewidth]{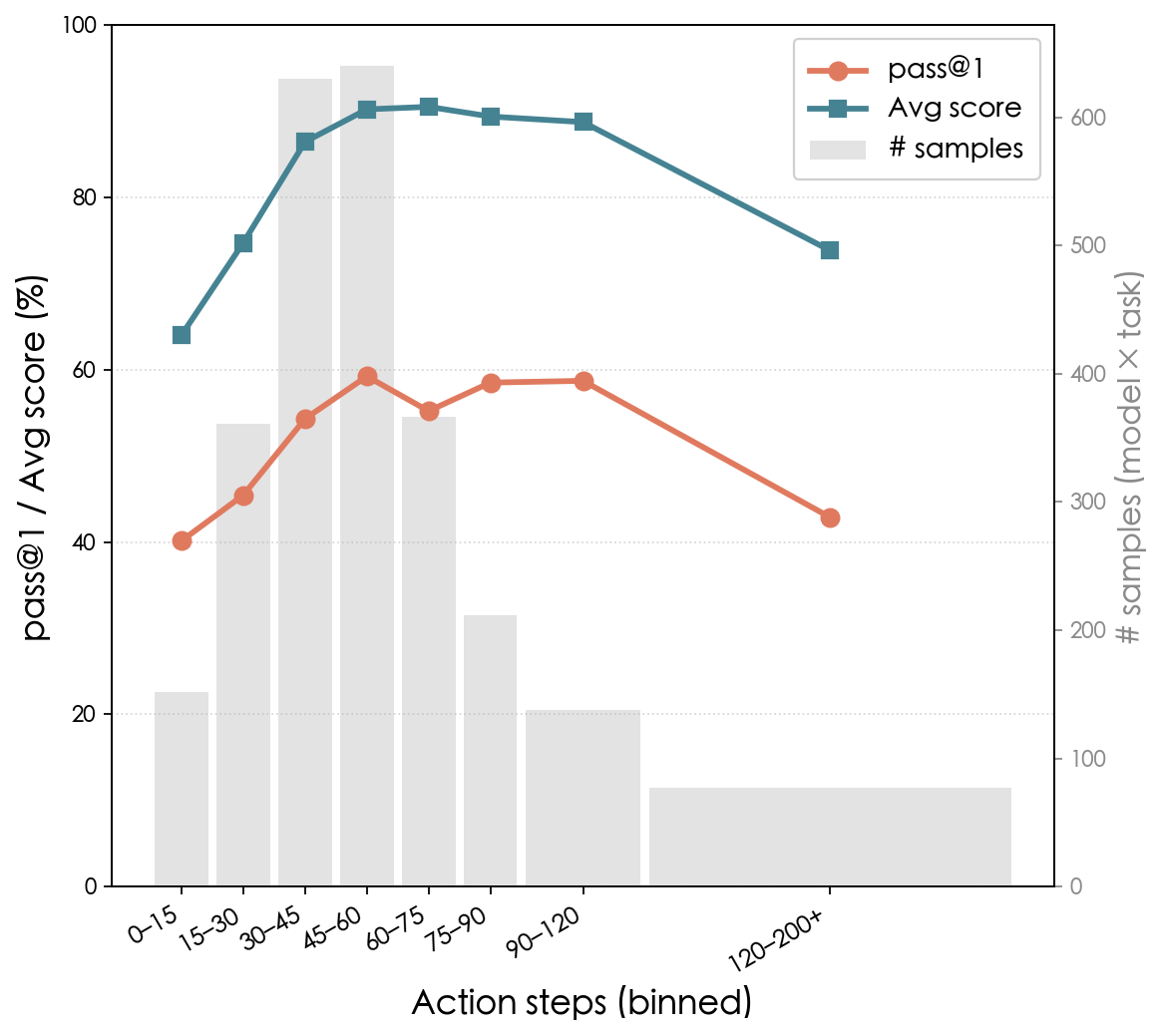}
        \caption{Outcome vs.\ effort across action-step bins.}
        \label{fig:outcome_effort}
    \end{subfigure}

    \caption{\textbf{Action-step efficiency and task complexity.}
    (a) Distribution of action steps per model on the commonly-solved set.
    (b) pass@1 and average score across action-step bins, with sample counts.}
    \label{fig:step-analysis}
\end{figure}

\textbf{OmniaBench remains challenging for current frontier models.}
As reported in Table~\ref{tab:benchmark}, even the frontier models Claude-Sonnet-5 and GPT-5.6-Sol, attains an Overall Pass@1 of only 58.54\% and 57.14\%, respectively, followed closely by GLM-5.2 and GPT-5.5 at 56.83\% and 56.52\%, respectively. The substantial performance spread across the remaining models further demonstrates the discriminative power of OmniaBench across different model families and capability levels. Figure~\ref{fig:score_distribution} shows that, even among the leading models, only approximately half of the tasks are fully solved, while a considerable proportion are either partially completed or failed, indicating that models frequently make meaningful progress without satisfying all task requirements. Consistently, the ToolSteps statistics in Table~\ref{tab:benchmark} and Figure~\ref{fig:step-analysis} show that evaluation trajectories often span dozens of tool calls and extended action sequences. These results highlight the sustained planning, information aggregation, state tracking, and cross-tool coordination required by OmniaBench.

\begin{figure}[t]
\vspace{-2em}
    \centering
    \includegraphics[
        width=0.98\linewidth
    ]{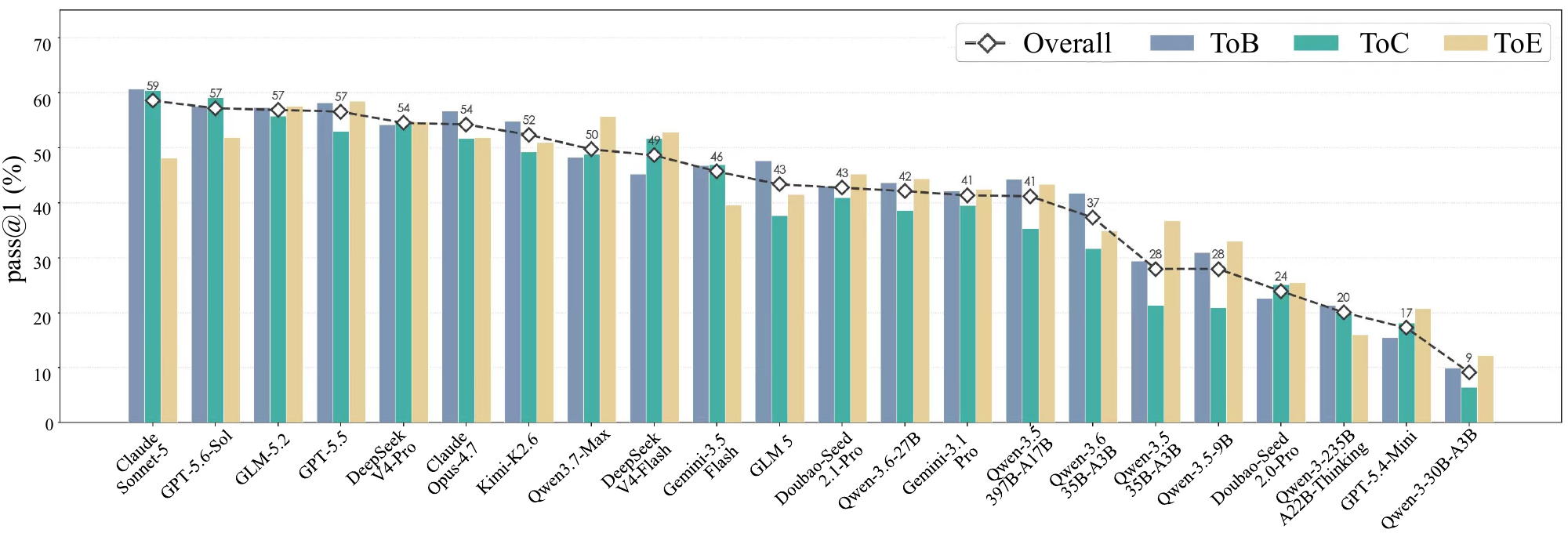}
    \vspace{-0.6em}
    \caption{\textbf{Performance across domain splits.}
    Pass@1 results on splits, with diamonds denoting overall scores.}
    \label{fig:scenario-split-performance}
    \vspace{-0.4em}
\end{figure}

\textbf{Models exhibit substantial performance variation across both capability and domains.}
As shown in Figure~\ref{fig:capability_radar}, performance across the ten capability dimensions is highly non-uniform: different models develop distinct strengths in Task Understanding, Information Gathering, Planning \& Decision Making, State Management, and Tool Use. Consequently, similar overall scores may correspond to markedly different capability profiles. Clear variation is also observed across domains. Figure~\ref{fig:scenario-split-performance} shows that Claude-Sonnet-5, GPT-5.6-Sol, and GLM-5.2 achieve ToB scores of 60.68\%, 57.59\%, and 57.28\%, respectively, substantially exceeding some of their results on the other domain splits, whereas GLM-5.2 attains its strongest split-level performance on ToE at 57.55\%. Thus, even models with comparable Overall scores may exhibit substantially different relative strengths across ToB, ToC, and ToE, suggesting that a single aggregate metric cannot fully characterize their practical applicability. Figure~\ref{fig:domain-performance} further corroborates this observation at the level-1 domain granularity, revealing additional shifts in model rankings across specific application scenarios.

\textbf{Comparable overall scores can mask large differences in tool-use efficiency.}
As shown in the left panel of Figure~\ref{fig:step-analysis}, models with similar overall success rates can differ substantially in the length and structure of their execution trajectories. Some models complete tasks through compact, well-targeted sequences of tool calls, whereas others require markedly longer and more circuitous interactions to reach comparable outcomes. Such longer trajectories often reflect redundant exploration, repeated or ineffective calls, unnecessary replanning, or recovery from execution errors rather than more thorough task completion. More efficient models identify the required information and operations more precisely, avoid unnecessary interactions, and satisfy task requirements with fewer steps. OmniaBench therefore captures not only whether a task is completed, but also how efficiently tools are planned and orchestrated.

\subsection{Discussion}

\begin{figure}[h]
    \centering

    \begin{subfigure}[b]{0.42\linewidth}
        \centering
        \includegraphics[width=\linewidth]{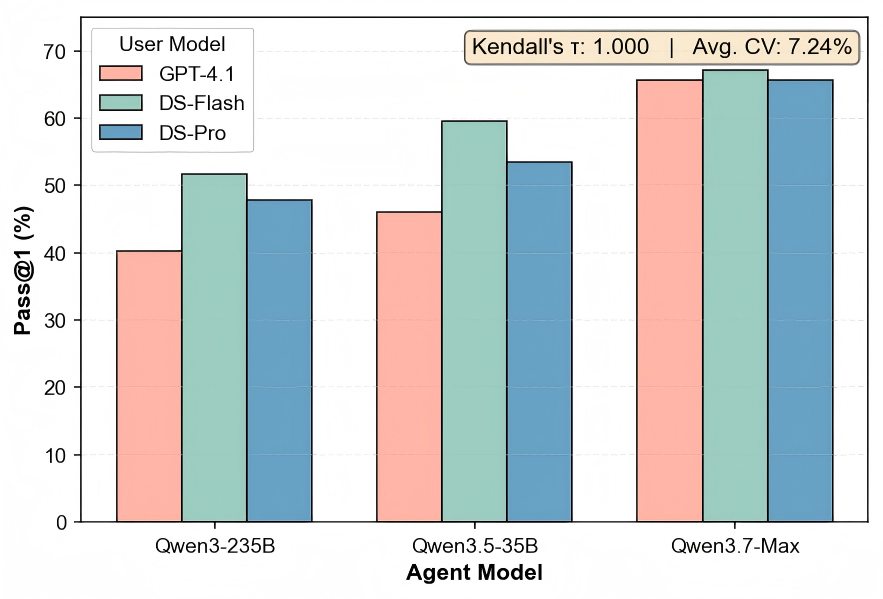}
        \caption{Robustness across user models.}
        \label{fig:user_robustness}
    \end{subfigure}
    \hspace{0.06\linewidth}
    \begin{subfigure}[b]{0.46\linewidth}
        \centering
        \raisebox{1em}{
        \includegraphics[width=\linewidth]{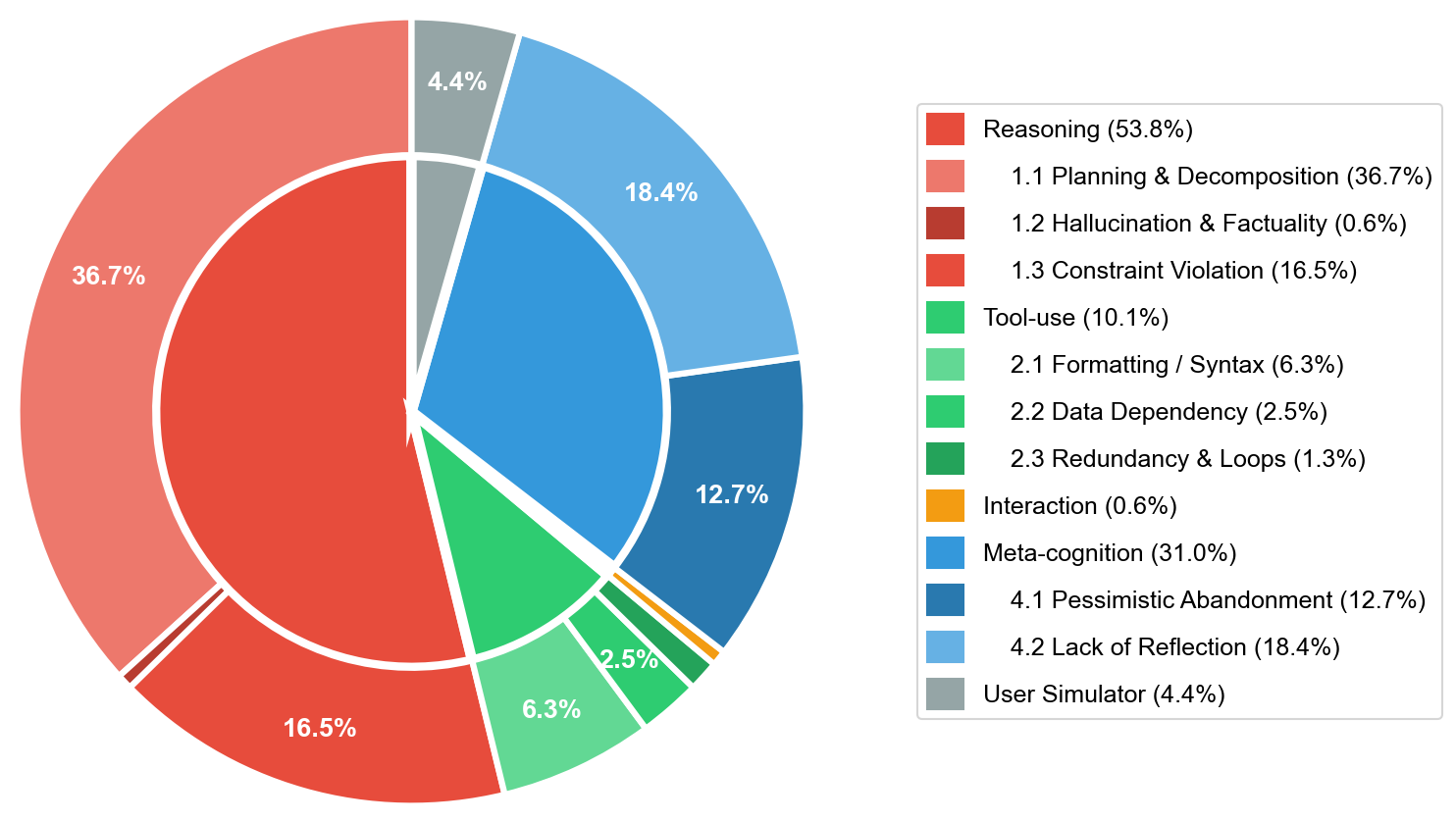}
        }
        \caption{Distribution of agent errors.}
        \label{fig:error_distribution}
    \end{subfigure}

    \caption{\textbf{Evaluation robustness and error analysis.}
    (a) Pass@1 results under different user simulators, showing stable model rankings.
    (b) Distribution of error categories observed during Claude-Opus-4.7 execution.}
    \label{fig:robustness_error_analysis}
    \vspace{-1em}
\end{figure}

\textbf{Reliability of User Simulation and Repeated Exploration.}
As shown in Figure~\ref{fig:user_robustness}, replacing the user simulator in multi-turn interactions with GPT-4.1, DeepSeek-V4-Flash, or DeepSeek-V4-Pro leaves the relative ranking of the three evaluated models unchanged, yielding a Kendall's \(\tau\) of 1.0 and an average coefficient of variation of 7.24\% across user models. Although different user simulators induce moderate shifts in absolute
\begin{wrapfigure}{r}{0.45\textwidth}
    \centering
    \vspace{-0.8\baselineskip}
    \includegraphics[width=\linewidth]{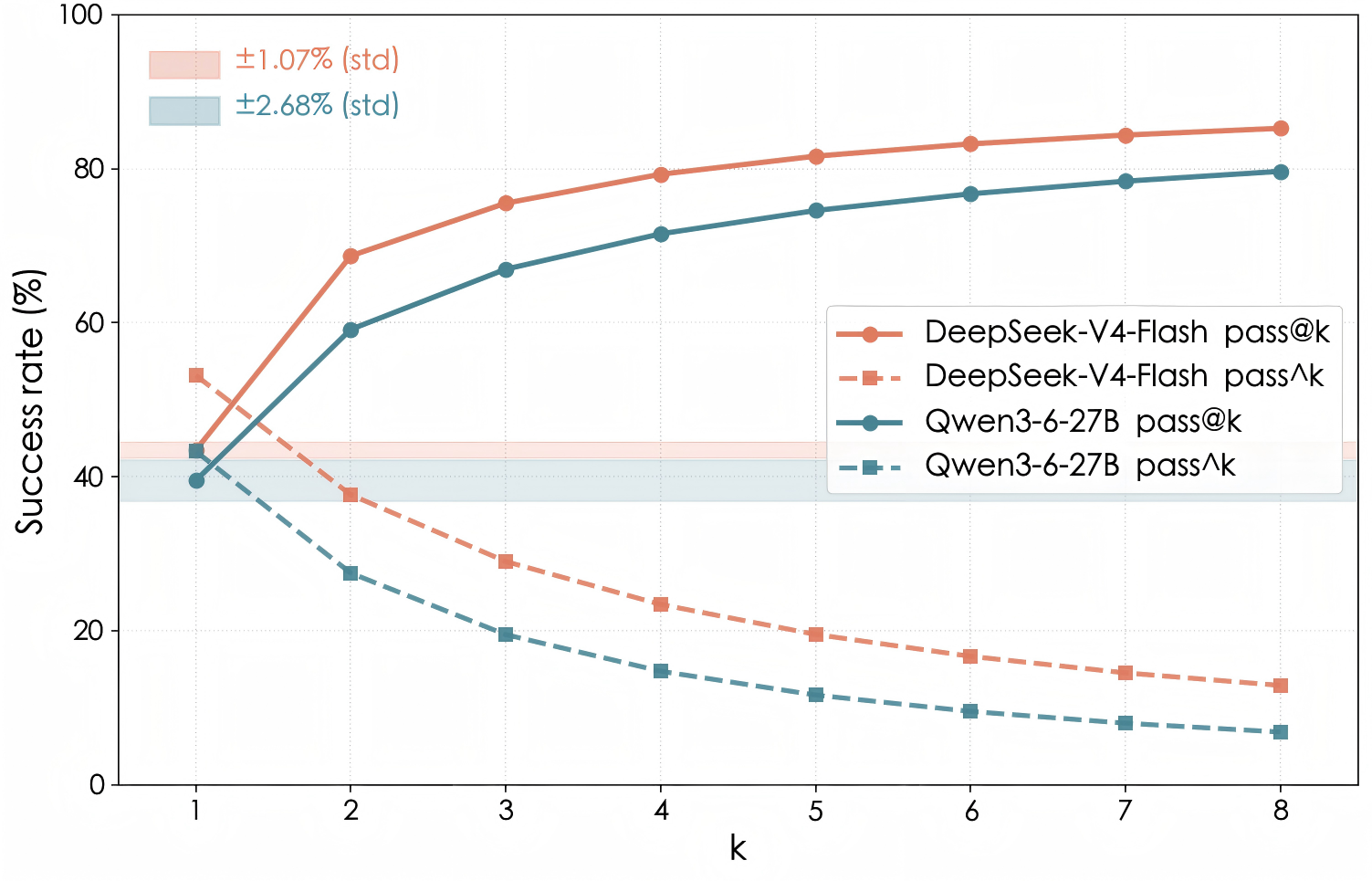}
    \caption{\textbf{Multi-run performance.}
    Pass@\(k\) and Pass\textsuperscript{\^{}}k results across two models.}
    \label{fig:passk-passpowk}
    \vspace{-0.6\baselineskip}
\end{wrapfigure}
scores, the principal comparative conclusions remain stable. Figure~\ref{fig:passk-passpowk} further shows that Pass@\(k\) increases with the number of independent runs, indicating that restarting from scratch provides additional opportunities to solve difficult tasks. At the same time, Pass\textsuperscript{\^{}}k declines rapidly, suggesting that such successes are not yet consistently reproducible across repeated executions. Repeated exploration therefore improves task coverage, but incurs additional execution cost and highlights the remaining gap toward reliable and efficient completion of complex tasks. The Pass@1 standard deviations of DeepSeek-V4-Flash and Qwen3.6-27B are only 1.07 and 2.68 percentage points, respectively, indicating that the aggregate results remain reproducible. We also conduct an independent stability analysis of the rubric judge and obtain consistent findings, as detailed in Appendix~\ref{appendix:judge-exp}.

\textbf{Distribution of Error Modes.}
As shown in Figure~\ref{fig:error_distribution}, reasoning-related failures account for 53.8\% of all observed errors, with planning and decomposition contributing 36.7\% and constraint violations a further 16.5\%. Meta-cognitive errors constitute another 31.0\%, primarily arising from insufficient reflection and premature abandonment. By comparison, direct tool-use errors, such as formatting mistakes, data-dependency failures, and redundant calls, account for a substantially smaller proportion. This distribution indicates that the main bottlenecks of current general-agent models lie less in basic tool invocation than in long-horizon planning, constraint maintenance, and adaptive correction based on execution feedback.


\textbf{Evaluation Priorities for Stronger General Agents.}
As overall task success continues to improve, a single aggregate score will become increasingly insufficient for characterizing differences among advanced models. Our results show that models with similar overall performance may still exhibit substantial variation across capability dimensions, domains, tool-use efficiency, and failure modes. Future evaluation should therefore place greater emphasis on structured analyses of specific abilities and domains, such as long-horizon planning, information integration, constraint adherence, and error recovery, rather than relying solely on task success. The domain, capability, and error-oriented analyses provided by OmniaBench offer a basis for identifying finer-grained capability boundaries among increasingly strong general agents.

\section{Conclusion}

We present OmniaBench, a general agent benchmark covering diverse application scenarios. OmniaBench builds executable environments from a hierarchical ToB, ToC, and ToE taxonomy and constructs tasks through multiple interaction routes. Experiments on frontier models demonstrate that OmniaBench remains challenging while revealing distinct capability profiles, domain specialization, tool-use efficiency, and failure patterns. OmniaBench provides a systematic foundation for evaluating and improving general agents.

\clearpage

\section*{Contributions}

The authors contributed to this work as follows.
The symbols $\ast$, $\dagger$, and $\ddagger$ denote equal contribution, project lead, and corresponding authors, respectively.
This work was conducted during the student authors' internships at Huawei.

\vspace{1em}

\noindent
\textbf{Chengyu Shen}$^{1,\ast}$,
\textbf{Yujie Fu}$^{2,\ast}$,
\textbf{Gangtao Xin}$^{3,\ast}$,
\textbf{Yanheng Hou}$^{1}$,
\textbf{Wenlong Fei}$^{4}$,
\textbf{Guojie Zhu}$^{3}$,
\textbf{Jiawei Li}$^{4}$,
\textbf{Hongcheng Gao}$^{5}$,
\textbf{Runming He}$^{1}$,
\textbf{Zhen Hao Wong}$^{1}$,
\textbf{Meiyi Qiang}$^{1}$,
\textbf{Hao Liang}$^{1,6}$,
\textbf{Zhao Cao}$^{2}$,
\textbf{Hao Jiang}$^{3,\dagger}$,
\textbf{Chong Chen}$^{3,\ddagger}$,
\textbf{Wentao Zhang}$^{1,6,\ddagger}$

\vspace{1em}

\noindent
$^{1}$ Peking University\\
$^{2}$ Renmin University of China\\
$^{3}$ Huawei Cloud Post-Training Team\\
$^{4}$ Beijing Institute of Technology\\
$^{5}$ Tsinghua University\\
$^{6}$ Zhongguancun Academy

\vspace{1em}

\noindent
Contact: \href{mailto:scuuy05@gmail.com}{scuuy05@gmail.com}

Correspondence: Chong Chen, Wentao Zhang

\section*{Acknowledgments}

We thank Yanyu Wu for providing valuable suggestions from a product perspective, 
including the initial classification strategies derived from app stores and subsequent discussions that helped refine the taxonomy design.

\clearpage

\bibliographystyle{plainnat-short}
\bibliography{main}

\clearpage

\beginappendix 

\section{Rubric-Based Judgment Robustness}
\subsection{Judge Model Robustness Results}
\label{appendix:judge-exp}
Since our benchmark relies on LLM-based judges to evaluate agent trajectories, we examine whether the resulting evaluation scores are overly sensitive to the choice of judge model. We re-evaluate the same set of trajectories using five different judge models under an identical rubric and judging protocol: \texttt{deepseek-v4-flash}, \texttt{gpt-4.1}, \texttt{deepseek-v4-pro}, \texttt{glm-5-thinking}, and \texttt{claude-opus-4-7-thinking}.

We evaluate judge-model robustness at both the trajectory and rubric-item levels. At the trajectory level, we measure absolute score agreement, ranking consistency, and normalized score deviation. At the rubric-item level, we examine agreement on point-valued item scores as well as agreement on normalized binary pass/fail decisions. Detailed metric definitions are provided in Appendix~\ref{app:judge_robustness_metrics}.

\begin{table*}[h]
\centering
\caption{
Robustness of benchmark scores under judge-model substitution.
}
\label{tab:judge_model_robustness}
\setlength{\tabcolsep}{8pt}
\renewcommand{\arraystretch}{1.12}
\small

\begin{adjustbox}{max width=\textwidth}
\begin{tabular}{lccccccc}
\toprule
\rowcolor{HeaderBlue}
\textbf{Level}
& \textbf{\#Units}
& \textbf{\#Judges}
& \textbf{ICC(A,1)}
& \textbf{Mean Spearman}
& \textbf{Min Spearman}
& \textbf{NMPD}
& \textbf{P95 NMPD} \\
\midrule

Total Score
& 1102
& 5
& 0.780
& 0.821
& 0.756
& 0.062
& 0.266 \\

Weighted Item Score
& 9808
& 5
& 0.741
& 0.856
& 0.811
& 0.007
& 0.060 \\

\bottomrule
\end{tabular}
\end{adjustbox}

\vspace{0.45em}
\begin{minipage}{0.95\textwidth}
\footnotesize
\noindent
\textit{Note.}
Total Score evaluates trajectory-level score robustness.
Weighted Item Score evaluates agreement on the original point-valued rubric-item scores.
For Weighted Item Score, ICC and Spearman are computed from the raw item scores, while NMPD is normalized by the maximum total score of the corresponding trajectory to quantify the local contribution of item-level judge disagreement.
\#Units denotes evaluation units for which all five judges provide valid scores.
NMPD denotes normalized mean pairwise deviation.
\end{minipage}

\end{table*}

\paragraph{Results.}
As shown in Table~\ref{tab:judge_model_robustness}, the trajectory-level scores are relatively stable across the evaluated judge models. The total-score $\mathrm{ICC}(A,1)$ is $0.780$, showing that the judge models assign broadly similar absolute scores to the same trajectories. The mean pairwise Spearman correlation is $0.821$, and the minimum pairwise correlation is $0.756$, indicating that the relative ranking of trajectories is largely preserved across judge models.

The total-score NMPD is $0.062$. In other words, the average absolute score difference between a pair of judge models corresponds to $6.2\%$ of the maximum score of the corresponding trajectory. The P95 NMPD is $0.266$, indicating that $95\%$ of trajectories have an average pairwise judge difference no greater than $26.6\%$ of their maximum score.

\begin{table*}[h]
\centering
\caption{
Agreement statistics for rubric-item pass/fail decisions.
}
\label{tab:item_decision_agreement}
\setlength{\tabcolsep}{7pt}
\renewcommand{\arraystretch}{1.12}
\small

\begin{adjustbox}{max width=\columnwidth}
\begin{tabular}{lcccccc}
\toprule
\rowcolor{HeaderBlue}
\textbf{Level}
& \textbf{\#Items}
& \textbf{\#Judges}
& \textbf{Fleiss' $\kappa$}
& \textbf{Full Agree}
& \textbf{One-off}
& \textbf{Split} \\
\midrule

Item Decision
& 9808
& 5
& 0.573
& 86.3\%
& 9.3\%
& 4.4\% \\

\bottomrule
\end{tabular}
\end{adjustbox}

\vspace{0.35em}
\begin{minipage}{0.95\columnwidth}
\footnotesize
\noindent
\textit{Note.}
Item scores are normalized by their point values into binary pass/fail decisions.
Full Agree denotes unanimous agreement among all judges.
One-off denotes a 4-vs-1 disagreement.
Split denotes a near-even 3-vs-2 disagreement.
\end{minipage}

\end{table*}

At the rubric-item decision level, Table~\ref{tab:item_decision_agreement} shows that $86.3\%$ of trajectory-item units receive unanimous pass/fail decisions from all five judges. A further $9.3\%$ exhibit a one-off $4$-vs-$1$ disagreement, while only $4.4\%$ exhibit a near-even $3$-vs-$2$ split.

Fleiss' $\kappa$ is $0.573$, indicating moderate chance-corrected agreement among the evaluated judge models. Together, these results show that most rubric items elicit consistent binary decisions across the five judge models under the fixed rubric and judging protocol. However, this agreement does not by itself establish the correctness of the judgments or rule out systematic biases shared by multiple judge models.

The point-valued rubric-item scores also exhibit substantial cross-judge consistency. The weighted-item-score $\mathrm{ICC}(A,1)$ is $0.741$, while the mean and minimum pairwise Spearman correlations are $0.856$ and $0.811$, respectively. These results indicate that the judges largely agree on both the absolute values and relative ordering of point-valued item scores.

The weighted-item-score NMPD is $0.007$. This means that, for an average trajectory-item unit, the absolute score difference between a pair of judge models corresponds to approximately $0.7\%$ of the maximum total score of the associated trajectory. The P95 NMPD is $0.060$, indicating that the corresponding local item-level difference is no greater than $6.0\%$ of the trajectory maximum score for $95\%$ of trajectory-item units.

This statistic measures the average local effect of disagreement on a single rubric item. It should not be interpreted as the cumulative effect of all item-level disagreements on a trajectory's final score; cumulative score variation is directly captured by the trajectory-level Total Score NMPD.

Overall, the results provide evidence that aggregate benchmark scores are reasonably stable across the five evaluated judge models under the fixed rubric and judging protocol. The benchmark exhibits relatively high trajectory-level score agreement and ranking consistency, while binary item-level decisions show moderate chance-corrected agreement.

\subsection{Judge Model Robustness Metrics}
\label{app:judge_robustness_metrics}

This appendix provides the formal definitions of the metrics reported in
Tables~\ref{tab:judge_model_robustness} and~\ref{tab:item_decision_agreement}.

\paragraph{Notation.}
Let $N$ denote the number of evaluated trajectories and $M$ denote the
number of judge models.
For trajectory $i\in\{1,\ldots,N\}$ and judge model
$j\in\{1,\ldots,M\}$, let $s_{ij}$ denote the total score assigned by
judge $j$, and let $P_i$ denote the maximum attainable total score of
trajectory $i$.

Each trajectory $i$ contains $L_i$ rubric items.
For item $\ell\in\{1,\ldots,L_i\}$, let $z_{i\ell j}$ denote the raw
item score assigned by judge $j$, and let $p_{i\ell}$ denote the maximum
point value of that item.

Each rubric item follows an all-or-nothing scoring rule: a judge assigns
either zero points or the full point value of the item, and partial credit
is not allowed. Therefore,
\begin{equation}
z_{i\ell j}\in\{0,p_{i\ell}\}.
\end{equation}
To remove differences in item point values, we normalize each item score
into a binary pass/fail decision:
\begin{equation}
d_{i\ell j}=
\frac{z_{i\ell j}}{p_{i\ell}}
\in\{0,1\},
\end{equation}
where $d_{i\ell j}=1$ indicates that judge $j$ considers item $\ell$
satisfied, and $d_{i\ell j}=0$ indicates that the item is not satisfied.

Unless otherwise specified, all metrics are computed using complete-case
evaluation units for which all $M$ judge models provide valid scores.

\paragraph{Generic evaluation matrix.}
For each score-level analysis, we construct a score matrix
\begin{equation}
X\in\mathbb{R}^{n\times M},
\end{equation}
where rows correspond to evaluation units and columns correspond to judge
models. Let $x_{ij}$ denote the score assigned to evaluation unit $i$ by
judge model $j$.

For trajectory-level analysis, each evaluation unit is one trajectory.
For rubric-item score analysis, each evaluation unit is one
trajectory-item pair.

\paragraph{Absolute-agreement ICC.}
We report the absolute-agreement intraclass correlation coefficient,
$\mathrm{ICC}(A,1)$, to measure whether different judge models assign
similar absolute scores to the same evaluation units.

Following the two-way random-effects, single-measure,
absolute-agreement formulation, we compute:
\begin{equation}
\mathrm{ICC}(A,1) = 
\frac{MS_R-MS_E}
{
MS_R
+
(M-1)MS_E
+
\frac{M(MS_C-MS_E)}{n}
},
\end{equation}
where $MS_R$, $MS_C$, and $MS_E$ denote the ANOVA mean squares associated
with evaluation units, judge models, and residual error, respectively.
A higher $\mathrm{ICC}(A,1)$ indicates stronger absolute agreement across
judge models.

\paragraph{Pairwise Spearman correlation.}
To measure ranking consistency, we compute the Spearman rank correlation
for every pair of judge models $(a,b)$:
\begin{equation}
\rho_{ab} = 
\operatorname{corr}
\left(
\operatorname{rank}(x_{\cdot a}),
\operatorname{rank}(x_{\cdot b})
\right).
\end{equation}

We report both the mean and minimum Spearman correlation across all
$\binom{M}{2}$ judge pairs:
\begin{equation}
\mathrm{MeanSpearman}=
\frac{1}{\binom{M}{2}}
\sum_{1\leq a<b\leq M}
\rho_{ab},
\end{equation}
and
\begin{equation}
\mathrm{MinSpearman} =
\min_{1\leq a<b\leq M}
\rho_{ab}.
\end{equation}

Mean Spearman reflects average ranking consistency across judge models,
whereas Min Spearman captures the weakest-performing judge pair.

\paragraph{Normalized mean pairwise deviation.}
To quantify the magnitude of score variation across judge models, we
compute the normalized mean pairwise deviation.

For each evaluation unit $i$, define:
\begin{equation}
\Delta_i=
\frac{
\frac{1}{\binom{M}{2}}
\sum_{1\leq a<b\leq M}
\left|x_{ia}-x_{ib}\right|
}
{q_i},
\end{equation}
where $q_i$ is an analysis-specific normalization denominator.

The normalized mean pairwise deviation is:
\begin{equation}
\mathrm{NMPD}=
\frac{1}{n}
\sum_{i=1}^{n}
\Delta_i.
\end{equation}

We additionally report the empirical 95th percentile of the
unit-level normalized deviations:
\begin{equation}
\mathrm{P95\ NMPD}
=
Q_{0.95}
\left(
\{\Delta_i\}_{i=1}^{n}
\right),
\end{equation}
where $Q_{0.95}$ denotes the empirical 95th-percentile operator.

NMPD measures the average normalized pairwise score difference, whereas
P95 NMPD characterizes instability in the upper tail of the evaluation
units.

\paragraph{Metric instantiations.}
We instantiate the generic score matrix for two score-level analyses:
trajectory-level total scores and point-valued rubric-item scores.
Binary rubric-item decisions are analyzed separately using Fleiss'
$\kappa$ and raw agreement-pattern statistics.

\subparagraph{Trajectory-level total score.}
For trajectory-level total scores, each evaluation unit corresponds to
one trajectory, and we define:
\begin{equation}
x_{ij}=s_{ij},
\qquad
q_i=P_i.
\end{equation}

The corresponding unit-level normalized deviation is:
\begin{equation}
\Delta_i^{\mathrm{total}}=
\frac{
\frac{1}{\binom{M}{2}}
\sum_{1\leq a<b\leq M}
|s_{ia}-s_{ib}|
}
{P_i}.
\end{equation}

Thus, total-score NMPD measures the average pairwise judge difference in
trajectory score, normalized by the maximum attainable score of that
trajectory.

\subparagraph{Point-valued rubric-item score.}
For point-valued rubric-item scores, each evaluation unit corresponds to
a trajectory-item pair $(i,\ell)$, and we define:
\begin{equation}
x_{i\ell j}=z_{i\ell j},
\qquad
q_{i\ell}=P_i.
\end{equation}

ICC and Spearman correlation are computed directly from the raw
point-valued rubric-item scores $z_{i\ell j}$. These scores jointly reflect
the binary pass/fail decisions made by the judges and the predefined point
values of the rubric items.

Let
\begin{equation}
\mathcal{U}=
\left\{
(i,\ell):
\text{all $M$ judges provide a valid score for item $\ell$ of trajectory $i$}
\right\}
\end{equation}
denote the complete-case set of valid trajectory-item units, and let
\begin{equation}
U=|\mathcal{U}|
\end{equation}
denote its cardinality.

The item-score NMPD is:
\begin{equation}
\mathrm{NMPD}_{\mathrm{item}}=
\frac{1}{U}
\sum_{(i,\ell)\in\mathcal{U}}
\frac{
\frac{1}{\binom{M}{2}}
\sum_{1\leq a<b\leq M}
|z_{i\ell a}-z_{i\ell b}|
}
{P_i}.
\end{equation}

The corresponding P95 NMPD is the empirical 95th percentile of:
\begin{equation}
\Delta_{i\ell}^{\mathrm{item}}=
\frac{
\frac{1}{\binom{M}{2}}
\sum_{1\leq a<b\leq M}
|z_{i\ell a}-z_{i\ell b}|
}
{P_i},
\qquad
(i,\ell)\in\mathcal{U}.
\end{equation}

This metric is a micro-average over trajectory-item units. Consequently,
a trajectory containing more rubric items contributes more units to the
metric. It measures the average local contribution of pairwise judge
disagreement on a single rubric-item score relative to the maximum total
score of the associated trajectory. It does not measure the cumulative
effect of all item-level disagreements within an entire trajectory.

\paragraph{Fleiss' $\kappa$ for binary item decisions.}
For normalized binary item decisions, we report Fleiss' $\kappa$ to
measure multi-judge agreement after correcting for agreement expected by
chance.

Each complete-case trajectory-item pair is treated as one evaluation unit
$u\in\{1,\ldots,U\}$. Let $n_{u,1}$ denote the number of judges assigning
a pass decision and let $n_{u,0}$ denote the number assigning a fail
decision. By definition,
\begin{equation}
n_{u,0}+n_{u,1}=M.
\end{equation}

The observed agreement for unit $u$ is:
\begin{equation}
P_u=
\frac{1}{M(M-1)}
\sum_{c\in\{0,1\}}
n_{u,c}\bigl(n_{u,c}-1\bigr).
\end{equation}

The mean observed agreement across all item-decision units is:
\begin{equation}
\bar{P}=
\frac{1}{U}
\sum_{u=1}^{U}
P_u.
\end{equation}

The empirical marginal proportion of category $c\in\{0,1\}$ is:
\begin{equation}
p_c=
\frac{1}{UM}
\sum_{u=1}^{U}
n_{u,c}.
\end{equation}

The expected agreement under the empirical marginal category
distribution is:
\begin{equation}
\bar{P}_e=
\sum_{c\in\{0,1\}}
p_c^2.
\end{equation}

Fleiss' $\kappa$ is then defined as:
\begin{equation}
\kappa=
\frac{\bar{P}-\bar{P}_e}
{1-\bar{P}_e}.
\end{equation}

A higher value of $\kappa$ indicates stronger chance-corrected agreement
among the judge models.

\paragraph{Agreement patterns.}
In addition to chance-corrected agreement, we report three raw
agreement-pattern statistics.

For each item-decision unit $u$, define:
\begin{equation}
r_u=
\max(n_{u,0},n_{u,1}),
\end{equation}
where $r_u$ is the number of judges supporting the majority decision.

The full-agreement ratio is:
\begin{equation}
\mathrm{FullAgree}=
\frac{1}{U}
\sum_{u=1}^{U}
\mathbf{1}[r_u=M].
\end{equation}
This measures the fraction of item-decision units for which all judge
models produce the same binary decision.

The one-off disagreement ratio is:
\begin{equation}
\mathrm{OneOff}=
\frac{1}{U}
\sum_{u=1}^{U}
\mathbf{1}[r_u=M-1].
\end{equation}
For $M=5$, this corresponds to a $4$-vs-$1$ disagreement.

The split ratio is:
\begin{equation}
\mathrm{Split}=
\frac{1}{U}
\sum_{u=1}^{U}
\mathbf{1}
\left[
\left\lceil\frac{M}{2}\right\rceil
\leq r_u
\leq M-2
\right].
\end{equation}
For $M=5$, this corresponds to a $3$-vs-$2$ split. Such cases represent
the strongest possible disagreement pattern among an odd number of five
judges and may indicate more ambiguous or difficult rubric-item
decisions.

\section{Taxonomy Details}
\label{appendix:taxonomy-details}

\subsection{Capability Dimensions}
\label{appendix:capability-taxonomy}

The capability taxonomy provides a unified diagnostic view of the behaviors required across the heterogeneous domains and data routes in OmniaBench. As described in Section~\ref{sec:data_construction}, the taxonomy is used to guide scenario selection, environment construction, task synthesis, and subsequent evaluation design. Rather than treating tool interaction as a single undifferentiated ability, we decompose general-agent competence into ten dimensions covering task interpretation, information acquisition, planning, execution, state tracking, artifact manipulation, interaction, and reliability.

A task may involve multiple capability dimensions simultaneously because
realistic agent tasks often require several interdependent behaviors.
We therefore associate each task with all applicable capability labels.
For capability-level aggregation, a task contributes to every capability
dimension with which it is annotated. The complete definitions of the ten
capability dimensions are provided in Table~\ref{tab:capability-axis}.

\begin{table}[h]
\centering
\small
\setlength{\tabcolsep}{4pt}
\renewcommand{\arraystretch}{1.10}
\caption{\textbf{Capability dimensions.}
Each task may be annotated with multiple capability dimensions and contributes to the aggregation of every dimension with which it is associated.}
\label{tab:capability-axis}
\begin{tabularx}{\linewidth}{@{}p{0.31\linewidth}X@{}}
\toprule
\textbf{Capability} & \textbf{What it evaluates} \\
\midrule

Task Understanding
& Identifying user goals, implicit requirements, priorities, domain constraints, and expected outcomes. \\

Information Gathering
& Locating, retrieving, and integrating relevant evidence from environment states, tools, files, databases, and external information sources. \\

Planning \& Decision Making
& Decomposing goals, selecting execution strategies, respecting dependencies and constraints, and revising plans as new observations become available. \\

State Management
& Tracking intermediate progress and maintaining consistency across entities, environment states, and long or multi-turn trajectories. \\

Tool Use
& Selecting appropriate tools, constructing valid arguments, interpreting outputs, and coordinating multiple tool calls. \\

Code \& Programmatic Operations
& Writing and executing code for computation, data transformation, automation, file manipulation, and programmatic task completion. \\

Data Analysis
& Filtering, aggregating, comparing, reconciling, and reasoning over structured or semi-structured data. \\

Office \& Document Handling
& Reading, extracting, editing, merging, validating, and producing documents, spreadsheets, presentations, and other file-based artifacts. \\

Interactive Collaboration
& Requesting missing information, clarifying ambiguous goals, confirming actions, incorporating user feedback, and coordinating across interaction turns. \\

Reliability \& Safety
& Detecting and recovering from failures, maintaining constraint compliance, avoiding unsafe or invalid actions, and completing tasks robustly under uncertainty. \\

\bottomrule
\end{tabularx}
\end{table}

\subsection{Atomic Difficulty Factors}
\label{appendix:atomic-difficulty}

In addition to capability coverage, we characterize task complexity using a set of atomic difficulty factors. Capability dimensions describe \emph{what} an agent is required to do, whereas difficulty factors describe \emph{why} the corresponding task is challenging. This separation allows tasks that evaluate the same primary capability to instantiate different sources of difficulty, such as ambiguous requirements, large structured states, long evidence contexts, dynamic dependencies, or inconsistent information.

The difficulty atoms are used as compositional design primitives during environment and task construction. A single task may combine multiple atoms, and the same atom may appear across different domains and data routes. For example, a DAG task may combine progressive information disclosure, state evolution, and dynamic multi-step planning, while a Solver task may combine structured-information complexity with conflicting constraints. These atoms are intended to support controlled task diversification and fine-grained error analysis rather than define mutually exclusive task categories. Table~\ref{tab:difficulty-atoms} summarizes the eight atomic difficulty factors and their typical instantiations.

\begin{table}[h]
\centering
\small
\setlength{\tabcolsep}{4pt}
\renewcommand{\arraystretch}{1.10}
\caption{\textbf{Atomic difficulty dimensions.} Multiple atoms may be composed within a single task.}
\label{tab:difficulty-atoms}
\begin{tabularx}{\linewidth}{@{}p{0.31\linewidth}X@{}}
\toprule
\textbf{Difficulty Atom} & \textbf{Instantiation} \\
\midrule

Ambiguous Goal and Contextual Constraints
& The request is underspecified, indirect, or conditioned on implicit preferences, policies, priorities, or professional constraints. \\

Tool and Parameter Grounding
& The agent must distinguish similar or redundant tools, infer arguments from context, or request missing parameters before execution. \\

Structured-Information Complexity
& The environment contains numerous structured entities, attributes, relations, or records that must be filtered, joined, compared, or reconciled. \\

Long-Context and Multi-Artifact Evidence
& Relevant evidence is distributed across long tool outputs, documents, files, attachments, logs, or multiple heterogeneous artifacts. \\

Dynamic Multi-Step Planning
& Completion requires long dependencies, conditional branches, intermediate decisions, or replanning after new observations. \\

Multi-Source Inconsistency
& Information from users, tools, files, or environment states is incomplete, duplicated, outdated, or mutually conflicting. \\

Progressive Disclosure and State Evolution
& Critical information or constraints are revealed gradually through user turns, tool results, approval stages, or state transitions. \\

Risk, Reliability, and Clarification
& The task involves irreversible actions, insufficient evidence, conflicting instructions, or operations that require explicit confirmation, recovery handling, or refusal. \\

\bottomrule
\end{tabularx}
\end{table}

\subsection{Representative Domain-Level Analysis}
\label{appendix:domain-analysis}


\begin{figure}[h]
    \centering
    \includegraphics[
        width=\linewidth
    ]{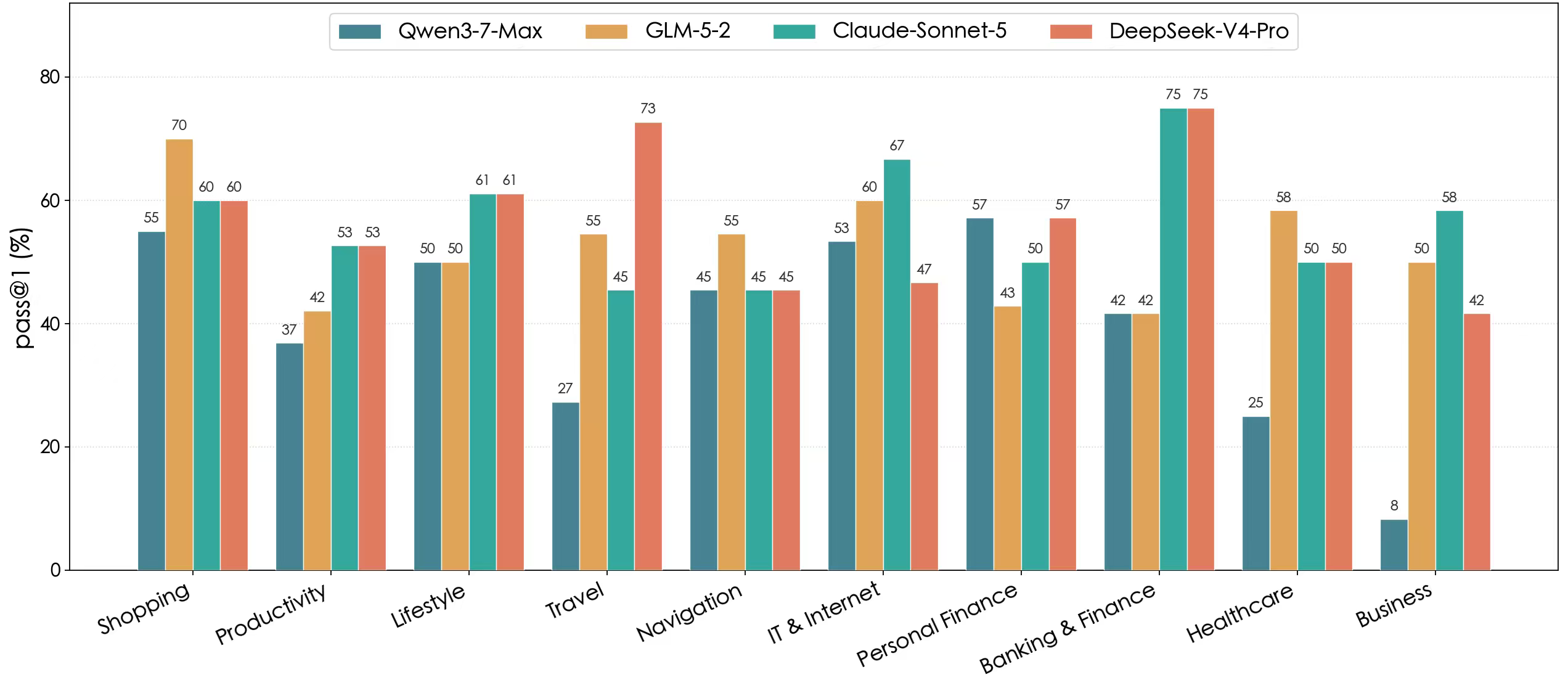}
    \caption{\textbf{Performance across representative agent scenarios.}
    Pass@1 results of four representative models over ten level-1 domains.}
    \label{fig:domain-performance}
\end{figure}

To complement the aggregate results across ToB, ToC, and ToE, we further compare four representative models over ten level-1 domains. As shown in Figure~\ref{fig:domain-performance}, model rankings vary substantially across application scenarios. GLM-5.2 performs best in Shopping and Healthcare, Claude-Opus-4.7 leads in Productivity, Lifestyle, and Navigation, while Qwen3.7-Max achieves the strongest result in IT \& Internet. DeepSeek-V4-Pro performs particularly well in Travel, Banking \& Finance, and Business, and matches Qwen3.7-Max on Personal Finance. These pronounced rank reversals indicate that no model consistently dominates across domains, and that similar aggregate scores may conceal substantially different scenario-specific strengths. The results further motivate evaluating general agents with broad domain coverage and reporting fine-grained performance beyond a single overall score.

\section{Full-Set Data Statistics}
\label{appendix:full-data-statistics}

Table~\ref{tab:multi_category_overall_statistics_appendix} summarizes the statistics of the full OmniaBench dataset, complementing the challenging-set statistics reported in the main text. The full set contains 1,431 tasks in total, including 1,102 DAG tasks, 222 DAG-S tasks, 60 Solver tasks, and 47 Program tasks. Although the four routes share a common environment and evaluation framework, their distinct construction objectives lead to substantially different task and environment characteristics.

\begin{table*}[h]
\centering
\caption{
Overall statistics of different task categories in full-set data.
All values are averaged over tasks within each category except \#Tasks. The Program route uses a binary \VerifyCode{} verifier instead of multi-item textual rubrics.
}
\label{tab:multi_category_overall_statistics_appendix}
\setlength{\tabcolsep}{9pt}
\renewcommand{\arraystretch}{1.12}
\small

\begin{adjustbox}{max width=\textwidth}
\begin{tabular}{lrrrrr}
\toprule
\rowcolor{HeaderBlue}
\textbf{Metric}
& \textbf{DAG}
& \textbf{DAG-S}
& \textbf{Solver}
& \textbf{Program}
& \textbf{Overall} \\
\midrule

\rowcolor{GroupGray}
\multicolumn{6}{l}{\textbf{Dataset Size}} \\
\#Tasks
& 1102
& 222
& 60
& 47
& \textbf{1431} \\

\addlinespace[0.25em]
\rowcolor{GroupGray}
\multicolumn{6}{l}{\textbf{Task}} \\
Task length
& 845.6
& 2933.7
& 1450.5
& 4431.4
& \textbf{1312.7} \\

\addlinespace[0.25em]
\rowcolor{GroupGray}
\multicolumn{6}{l}{\textbf{Environment}} \\
\#Tools
& 82.6
& 56.4
& 8.2
& 51.7
& \textbf{68.6} \\
\#Entity tables
& 16.4
& 13.1
& 6.7
& 12.9
& \textbf{14.9} \\
\#Entities
& 70.7
& 208.1
& 111.0
& 308.2
& \textbf{100.1} \\

\addlinespace[0.25em]
\rowcolor{GroupGray}
\multicolumn{6}{l}{\textbf{Rubric or VerifyCode}} \\
\#Items
& 8.9
& 13.0
& 5.8
& 1
& \textbf{9.4} \\
Item points
& 21.6
& 32.4
& 5.8
& 1
& \textbf{22.6} \\
Description length
& 117.1
& 252.8
& 429.3
& \VerifyCode{}
& \textbf{152.4} \\

\bottomrule
\end{tabular}
\end{adjustbox}


\end{table*}

Program tasks have the longest average task descriptions, reaching 4,431.4 tokens, reflecting their richer procedural constraints, branching structures, and execution requirements. DAG-S tasks are also relatively long, with an average length of 2,933.7, whereas DAG tasks are more concise on average. The routes additionally differ in environment scale. DAG tasks expose the largest tool space, with 82.6 candidate tools per task on average. In contrast, Program and DAG-S tasks contain denser state spaces, with 308.2 and 208.1 instantiated entities, respectively. Solver tasks involve fewer tools and entity tables, as their complexity primarily arises from structured constraints and optimization objectives rather than broad tool exploration.

The rubric statistics further reflect the differences among the four routes. DAG-S tasks contain the largest number of rubric items and rubric points, with averages of 13.0 and 32.4, respectively, indicating more fine-grained completion criteria. Solver tasks contain fewer rubric items, but their rubric descriptions are the longest on average, suggesting that their evaluation criteria require more detailed specification of feasibility, constraint satisfaction, and objective consistency. Across the full dataset except the program route, each task contains an average of 9.4 rubric items and 22.6 rubric points, providing structured and fine-grained supervision across diverse task formats.

\section{Route-Specific Evaluation Protocols}
\label{app:route-evaluation}

OmniaBench contains four data routes with different interaction and verification protocols. The DAG route adopts multi-turn interaction, whereas Solver, Program, and DAG-S are evaluated in single-turn settings.

\subsection{DAG: Multi-turn Interactive Evaluation}

For DAG tasks, a persona-grounded user simulator initiates the conversation and progressively provides task-relevant information according to the interaction history. At each turn, the evaluated agent may respond to the user, invoke tools, or request clarification. The interaction terminates when the user simulator emits \texttt{\#\#\#STOP\#\#\#}, or when the predefined execution budget is exhausted.

\begin{tcolorbox}[
    enhanced,
    breakable,
    colback=HeaderBlue!35,
    colframe=RuleGray,
    boxrule=0.5pt,
    arc=1mm,
    title={DAG Evaluation Flow},
    fonttitle=\bfseries
]
\begin{enumerate}
    \item The user simulator generates the initial request from the task specification and persona.
    \item The agent interacts with the user and environment through messages and tool calls.
    \item The user simulator reveals additional information or provides feedback when required.
    \item Multiple rubric items assess whether all task objectives and constraints are satisfied.
\end{enumerate}
\end{tcolorbox}

This protocol evaluates not only task execution, but also clarification, constraint tracking, adaptation to user feedback, and state maintenance across multiple turns.

\subsection{Solver and DAG-S: Single-turn Rubric Evaluation}

Solver and DAG-S tasks provide the full user request at the beginning of the trajectory. The agent may invoke multiple tools, but no additional user feedback is introduced after the initial request. Execution terminates when the agent submits its final response.

Both routes are evaluated using multiple rubric items. Each item corresponds to a concrete task objective or constraint and receives either its assigned score or zero. A task is considered successful only when all rubric items are satisfied. Solver tasks additionally emphasize constraint satisfaction and the consistency of the produced decision with the underlying planning, allocation, scheduling, or optimization objective.

\subsection{Program: Single-turn Programmatic Verification}

Program tasks also follow a single-turn protocol, but use programmatic verification instead of rubric-based judgment. After completing the required tool interactions, the agent submits its final response. The associated \VerifyCode{} then determines the binary result directly from the complete trajectory and final observation.

\begin{tcolorbox}[
    enhanced,
    breakable,
    colback=OverallGray,
    colframe=RuleGray,
    boxrule=0.5pt,
    arc=1mm,
    title={Program Verification Flow},
    fonttitle=\bfseries
]
\begin{enumerate}
    \item The agent receives the task request and available tool interfaces.
    \item The agent gathers information and performs the required operations.
    \item The agent submits its final response.
    \item \VerifyCode{} checks the trajectory and final observation and returns a binary pass/fail result.
\end{enumerate}
\end{tcolorbox}

\section{Persona Design}
\label{app:persona-design}

Personas are used in DAG tasks to create diverse and realistic user behaviors. Each persona specifies three groups of attributes:

\begin{itemize}
    \item \textbf{Core identity}: professional role, background, and life stage;
    \item \textbf{Interaction characteristics}: expression style, emotional tendency, information-disclosure strategy, decision preference, and patience;
    \item \textbf{Real-world constraints}: time, budget, privacy, and other scenario-dependent considerations.
\end{itemize}

These attributes control how the user expresses requests, reveals missing information, responds to clarification, and determines whether the interaction should continue.

\subsection{Persona Example: School Alumna}

\begin{tcolorbox}[
    enhanced,
    breakable,
    colback=OverallGray,
    colframe=RuleGray,
    boxrule=0.5pt,
    arc=1mm,
    title={Saint Saviour School Alumna},
    fonttitle=\bfseries
]
\textbf{Core identity.}
A graduate familiar with the school's academic culture, learning pace, and campus traditions.

\medskip
\textbf{Interaction characteristics.}
She communicates in an organized and experience-grounded manner, actively corrects inaccurate descriptions, and prefers focused questions. She is willing to discuss representative school experiences while withholding unnecessary personal details.

\medskip
\textbf{Constraints and preferences.}
She values long-term educational impact over superficial reputation, avoids disclosing information about other individuals, and expects the conversation to respect the school's specific cultural and educational context.
\end{tcolorbox}

\subsection{Persona Example: Television Producer}

\begin{tcolorbox}[
    enhanced,
    breakable,
    colback=OverallGray,
    colframe=RuleGray,
    boxrule=0.5pt,
    arc=1mm,
    title={Visionary Television Producer},
    fonttitle=\bfseries
]
\textbf{Core identity.}
An experienced television producer interested in socially meaningful and underrepresented stories.

\medskip
\textbf{Interaction characteristics.}
The producer frames discussions around characters, conflicts, themes, and narrative structure. They engage deeply with insightful ideas but quickly reject generic suggestions or formulaic expressions.

\medskip
\textbf{Constraints and preferences.}
They balance authenticity, communication value, filming feasibility, production cost, and ethical boundaries. Information about unaired projects and real subjects is disclosed cautiously.
\end{tcolorbox}

\section{User Prompt Composition}
\label{app:user-prompt}

The initial prompt provided to the user simulator combines three components:

\begin{equation}
\textsc{User Prompt}
=
\textsc{Persona}
\oplus
\textsc{Environment Summary}
\oplus
\textsc{Task Description}.
\end{equation}

\begin{tcolorbox}[
    enhanced,
    colback=HeaderBlue!35,
    colframe=RuleGray,
    boxrule=0.5pt,
    arc=1mm,
    title={User Prompt Template},
    fonttitle=\bfseries
]
\centering
\texttt{[PERSONA CARD]}
\quad+\quad
\texttt{[ENVIRONMENT SUMMARY]}
\quad+\quad
\texttt{[TASK DESCRIPTION]}
\end{tcolorbox}

\textbf{Persona card.}
The persona card determines the simulated user's background, communication pattern, information-disclosure behavior, preferences, and practical constraints.

\textbf{Environment summary.}
The environment summary briefly describes the application context represented by the available tools and state, such as a geographic information system connected to an asset-management database.

\textbf{Task description.}
The task description specifies the user objective, operational constraints, expected outcome, and any information that may be disclosed progressively during interaction. It is written from the user's perspective and avoids exposing the intended tool sequence or reference trajectory.

For example, a maintenance-dispatch task may ask the agent to identify at most one suitable work order while jointly considering urgency, asset availability, ongoing repair status, and recent assignment changes. The agent must retrieve the relevant evidence, resolve competing constraints, and avoid unnecessary state modifications.





\newpage
\definecolor{CodeTextGray}{RGB}{63,72,82}
\newtcolorbox{routeonecase}[1][]{%
  enhanced,
  breakable,
  colback=white,
  colframe=RuleGray,
  boxrule=0.75pt,
  arc=1.2mm,
  left=4mm,
  right=4mm,
  top=3mm,
  bottom=3mm,
  title={#1},
  colbacktitle=HeaderGray,
  coltitle=black,
  fonttitle=\bfseries,
  attach boxed title to top left={yshift=-2mm, xshift=4mm},
  boxed title style={
    colframe=RuleGray,
    colback=HeaderGray,
    boxrule=0.75pt,
    arc=1mm
  }
}

\newtcolorbox{routegroupbox}[1][]{%
  enhanced,
  breakable,
  colback=white,
  colframe=RuleGray,
  boxrule=0.5pt,
  arc=0.8mm,
  left=2.5mm,
  right=2.5mm,
  top=1.8mm,
  bottom=1.8mm,
  title={\small\bfseries #1},
  colbacktitle=GroupGray,
  coltitle=black
}

\lstdefinestyle{routejsonstylemain}{%
  basicstyle=\ttfamily\footnotesize\color{CodeTextGray},
  backgroundcolor=\color{HeaderGray},
  frame=single,
  framerule=0.4pt,
  rulecolor=\color{RuleGray},
  breaklines=true,
  columns=fullflexible,
  keepspaces=true,
  showstringspaces=false,
  xleftmargin=2mm,
  xrightmargin=2mm,
  aboveskip=4pt,
  belowskip=2pt
}

\section{Representative Cases}
\label{sec:route1-representative-cases}

We next present two representative route1 tasks. For each case, we show both an environment-entity view and a candidate-tool view so that the task can be read from two complementary angles: the structured state that the agent must reason over and the representative operations that the task invokes.

\subsection{Case 1}
\label{sec:route1-entity-case}

\begin{routeonecase}[Case A: Litigation Matter Environment]

\textbf{Environment:} \texttt{LawFirmLitigationMatterManagementSystem}
\hfill
\textbf{Global ID:} \texttt{803}\\
\hfill
\textbf{Domain:} \texttt{Legal Industry $\rightarrow$ Attorney}
\hfill 
\textbf{Task ID:} \texttt{env\_22\_task\_001}

\smallskip
\noindent\textbf{Task prompt (condensed).}
\begin{tcolorbox}[
  enhanced,
  colback=FoundationBlue,
  colframe=RuleGray,
  boxrule=0.5pt,
  arc=0.8mm,
  left=2mm,
  right=2mm,
  top=1.2mm,
  bottom=1.2mm
]
The task verifies the personal injury defense matter \texttt{LF-2024-PI-024} for Yunhe Foods. It confirms the client, responsible attorney, and opposing party, checks whether the archived investigation memorandum remains usable for linkage, reviews the evidence chain for completeness, and then adds a newly collected evidence item while updating the corresponding matter timeline.
\end{tcolorbox}

\smallskip
\noindent\textbf{Environment Entities snapshot.}

\noindent
\begin{minipage}[t]{0.35\textwidth}
\begin{routegroupbox}[Entity Groups]
  \textbf{Matter Core:}\\
  \texttt{matters}\\
  \texttt{clients}\\
  \texttt{attorneys}\\
  \textbf{Document Layer:}\\
  \texttt{documents}\\
  \texttt{document\_versions}
\end{routegroupbox}
\end{minipage}\hfill
\begin{minipage}[t]{0.63\textwidth}
\begin{routegroupbox}[Concrete Instance Slice]
\textbf{Matter profile}
\par\smallskip
\begin{tabularx}{\linewidth}{>{\raggedright\arraybackslash}p{0.37\linewidth}X}
\texttt{matter\_number} & \texttt{LF-2024-PI-024} \\
\texttt{client} & \texttt{Yunhe Foods Co.} \\
\texttt{attorney} & \texttt{Lin Qiaoxue} 
\end{tabularx}

\textbf{Evidence-chain slice}
\par\smallskip
\begin{tabularx}{\linewidth}{>{\raggedright\arraybackslash}p{0.37\linewidth}X}
\texttt{existing\_evidence} & \texttt{EVD-004} \\
\texttt{link\_id} & \texttt{LNK-003} 
\end{tabularx}
\end{routegroupbox}
\end{minipage}

\smallskip
\noindent\textbf{Candidate Tools snapshot.}

\noindent
\begin{minipage}[t]{0.35\textwidth}
\begin{routegroupbox}[Candidate Tool Groups]
  \textbf{Matter Resolution:}\\
  \texttt{get\_client\_by\_id}\\
  \texttt{get\_attorney\_by\_id}\\
  \textbf{Document \& Version Checks:}\\
  \texttt{get\_document}\\
  \texttt{list\_document\_versions}
\end{routegroupbox}
\end{minipage}\hfill
\begin{minipage}[t]{0.63\textwidth}
\begin{routegroupbox}[Concrete Tool Slice]
\textbf{Read-side checks}
\par\smallskip
\begin{tabularx}{\linewidth}{>{\raggedright\arraybackslash}p{0.40\linewidth}X}
\texttt{get\_client\_by\_id} & \texttt{CLT-002} \\
\texttt{get\_attorney\_by\_id} & \texttt{ATT-002} \\
\end{tabularx}

\textbf{Write-side actions}
\par\smallskip
\begin{tabularx}{\linewidth}{>{\raggedright\arraybackslash}p{0.40\linewidth}X}
\texttt{create\_evidence} & store surveillance screenshots \\
\texttt{create\_matter\_event} & supplement filing event \\
\end{tabularx}
\end{routegroupbox}
\end{minipage}

\smallskip
\noindent\textbf{Selected rubric criteria.}
\par\medskip
\noindent
\begin{tabularx}{\textwidth}{>{\raggedright\arraybackslash}p{0.12\textwidth} X}
\rowcolor{HeaderGray}
\textbf{Rubric} & \textbf{Criterion excerpt} \\
\midrule
\texttt{G1} & Correctly identify the target matter as the Yunhe Foods personal injury defense matter with matter number \texttt{LF-2024-PI-024}. \\
\texttt{T1} & Perform an integrity/completeness review for the matter and detect the evidence-chain problem, specifically the self-referential evidence-link anomaly. \\
\texttt{T4} & Create a corresponding matter event for the supplement and ensure the matter activity timeline reflects the addition without changing the matter's stayed status. \\
\end{tabularx}

\end{routeonecase}

\subsection{Case 2}

\begin{routeonecase}[Case B: Password Vault Environment]

\textbf{Environment:} \texttt{PasswordVaultAppEnvironment}
\hfill
\textbf{Global ID:} \texttt{825}\\
\hfill
\textbf{Domain:} \texttt{Tools $\rightarrow$ Password Management}
\hfill
\textbf{Task ID:} \texttt{env\_3\_task\_001}

\smallskip
\noindent\textbf{Task prompt (condensed).}
\begin{tcolorbox}[
  enhanced,
  colback=AgentOrange,
  colframe=RuleGray,
  boxrule=0.5pt,
  arc=0.8mm,
  left=2mm,
  right=2mm,
  top=1.2mm,
  bottom=1.2mm
]
The task performs a security cleanup for user \texttt{user\_zheng\_2048}: confirm accessible vaults and trusted devices, inspect a shared work login with unresolved breach risk, create a new work login and an encrypted note, verify passkey and note autofill behavior, and summarize export, reminder, and alert state across the vault.
\end{tcolorbox}

\smallskip
\noindent\textbf{Environment Entities snapshot.}

\noindent
\begin{minipage}[t]{0.37\textwidth}
\begin{routegroupbox}[Entity Groups]
  \textbf{Vault State:}\\
  \texttt{user\_vaults}\\
  \texttt{vault\_items}\\
  \texttt{categories}\\
  \textbf{Device Surface:}\\
  \texttt{device\_permissions}\\
  \texttt{sync\_states}\\
  \textbf{Credential Surface:}\\
  \texttt{login\_credentials}\\
  \texttt{passkey\_credentials}\\
  \textbf{Security \& Audit:}\\
  \texttt{share\_records}\\
  \texttt{export\_jobs}\\
  \texttt{import\_batches}
\end{routegroupbox}
\end{minipage}\hfill
\begin{minipage}[t]{0.61\textwidth}
\begin{routegroupbox}[Concrete Entity Slice]
\textbf{Vault and device state}
\par\smallskip
\begin{tabularx}{\linewidth}{>{\raggedright\arraybackslash}p{0.40\linewidth}X}
\texttt{vault\_id} & \texttt{vault\_main\_001} \\
\texttt{vault\_status} & \texttt{active} \\
\texttt{device\_id} & \texttt{DEV-ALPHA-01} \\
\texttt{device\_name} & \texttt{LunaBook Pro} \\
\texttt{is\_trusted} & \texttt{True} \\
\texttt{is\_online} & \texttt{True} \\
\end{tabularx}

\vspace{1mm}
\textbf{Credential and alert state}
\par\smallskip
\begin{tabularx}{\linewidth}{>{\raggedright\arraybackslash}p{0.40\linewidth}X}
\texttt{login\_item} & \texttt{ITEM\_LOGIN\_001} \\
\texttt{passkey\_item} & \texttt{ITEM\_PASSKEY\_002} \\
\texttt{secure\_note} & \texttt{ITEM\_NOTE\_004} \\
\texttt{alert\_id} & \texttt{ALERT\_001} \\
\texttt{share\_id} & \texttt{SHR\_001} \\
\texttt{export\_id} & \texttt{EXP\_001} \\
\end{tabularx}
\end{routegroupbox}
\end{minipage}

\smallskip
\noindent\textbf{Candidate Tools snapshot.}

\noindent
\begin{minipage}[t]{0.37\textwidth}
\begin{routegroupbox}[Candidate Tool Groups]
  \textbf{Vault \& Device Lookup:}\\
  \texttt{get\_vaults}\\
  \texttt{list\_devices}\\
  \texttt{get\_device\_permissions}\\
  \textbf{Credential Inspection:}\\
  \texttt{get\_login\_item}\\
  \texttt{get\_passkey\_item}\\
  \texttt{get\_secure\_note}\\
  \textbf{State-Changing Writes:}\\
  \texttt{create\_credential}\\
  \texttt{create\_secure\_note}\\
  \textbf{Security \& Audit Checks:}\\
  \texttt{list\_reminders}\\
  \texttt{get\_export\_job}\\
  \texttt{list\_share\_records}\\
  \texttt{get\_change\_history}
\end{routegroupbox}
\end{minipage}\hfill
\begin{minipage}[t]{0.61\textwidth}
\begin{routegroupbox}[Concrete Tool Slice]
\textbf{Read-side checks}
\par\smallskip
\begin{tabularx}{\linewidth}{>{\raggedright\arraybackslash}p{0.40\linewidth}X}
\texttt{get\_vaults} & \texttt{vault\_main\_001} \\
\texttt{list\_devices} & \texttt{DEV-ALPHA-01, DEV-BETA-02, DEV-GAMMA-03} \\
\texttt{get\_login\_item} & \texttt{ITEM\_LOGIN\_001} \\
\texttt{get\_passkey\_item} & \texttt{ITEM\_PASSKEY\_002} \\
\texttt{get\_secure\_note} & \texttt{ITEM\_NOTE\_004} \\
\texttt{list\_alerts} & \texttt{ALERT\_001 (open)} \\
\end{tabularx}

\vspace{1mm}
\textbf{Write-side actions and audit}
\par\smallskip
\begin{tabularx}{\linewidth}{>{\raggedright\arraybackslash}p{0.40\linewidth}X}
\texttt{create\_credential} & \texttt{https://hr.atlas.example} \\
\texttt{create\_secure\_note} & incident follow-up note \\
\texttt{list\_share\_records} & \texttt{SHR\_001} \\
\texttt{get\_export\_job} & \texttt{EXP\_001} \\
\texttt{get\_change\_history} & new-login trace required \\
\texttt{list\_reminders} & rotation / expiry reminders \\
\end{tabularx}
\end{routegroupbox}
\end{minipage}

\smallskip
\noindent\textbf{Selected rubric criteria.}
\par\medskip
\noindent
\begin{tabularx}{\textwidth}{>{\raggedright\arraybackslash}p{0.12\textwidth} X}
\rowcolor{HeaderGray}
\textbf{Rubric} & \textbf{Criterion excerpt} \\
\midrule
\texttt{G1} & Correctly identify the active vault owned by \texttt{user\_zheng\_2048} and the device best suited for sensitive actions as the trusted online laptop. \\
\texttt{G3} & Correctly summarize the vault security posture relevant to the review, including the compromised login, open alert, and due reminder horizon. \\
\texttt{T1--T2} & For the Aurora Mail login, verify lookup consistency, sharing state, and unresolved breach-alert status without inventing duplicate records. \\
\texttt{T5} & Aggregate export, import, reminder, and open-alert state and explicitly point out remaining inconsistencies or items needing human confirmation. \\
\end{tabularx}

\end{routeonecase}

\newpage
\section{Trajectory Case Study}
\subsection{Case 1}
\par\medskip
\noindent
\providecommand{\caseusertag}{}
\renewcommand{\caseusertag}{%
  \fcolorbox{RoleUser}{FoundationBlue}{\scriptsize\textsf{\textcolor{RoleUser}{User}}}%
}

\providecommand{\caseassistanttag}{}
\renewcommand{\caseassistanttag}{%
  \fcolorbox{RoleAssistant}{ScaleGreen}{\scriptsize\textsf{\textcolor{RoleAssistant}{Agent}}}%
}

\providecommand{\casetooltag}{}
\renewcommand{\casetooltag}{%
  \fcolorbox{RoleTool}{AgentOrange}{\scriptsize\textsf{\textcolor{RoleTool}{Tool}}}%
}
\small
\begin{tcolorbox}[
  enhanced,
  breakable,
  colback=white,
  colframe=RuleGray,
  boxrule=0.6pt,
  arc=1.5pt,
  title={\textbf{Case Example: Litigation Matter Evidence Update}},
  colbacktitle=HeaderGray,
  coltitle=black,
  fonttitle=\bfseries\small,
  left=4pt, right=4pt, top=3pt, bottom=3pt
]
\scriptsize
\textbf{\color{ActionBlue}[1] Task Description} \hfill \textit{task excerpt}\\[1pt]
Review litigation matter \texttt{LF-2024-PI-024} for Yunhe Foods: confirm client, responsible attorney, and opposing party; verify that the filed incident-investigation memo remains the current linkable version; run a completeness review with emphasis on evidence-chain anomalies; then add a new evidence item linked to \texttt{DOC-003}, and confirm the resulting trace record, activity timeline \ldots \\[2pt]
{\color{MutedText}\textit{[\ldots additional description omitted \ldots]}}

\tcbline

\textbf{\color{ActionBlue}[2] Tools} \hfill \textit{tool summary}\\[1pt]
Matter lookup by number; client and attorney retrieval; full matter workspace lookup; document and version retrieval; matter-integrity validation; evidence-subgraph and neighbor traversal; evidence insertion; matter-event lookup; activity-timeline retrieval.\\[2pt]
{\color{MutedText}\textit{[\ldots additional tools omitted \ldots]}}

\tcbline

\textbf{\color{ActionBlue}[3] Agent Setup} \hfill \textit{prompt excerpts}\\[1pt]
\begin{tabularx}{\textwidth}{@{}>{\bfseries}p{0.17\textwidth}X@{}}
Tested agent & ``You are a helpful assistant. Your goal is to complete the user's request in an interactive environment by gradually calling the available tools step by step, and to proactively communicate with the user \ldots \\
User agent & ``You are role-playing as a user interacting with an agent. Your persona is written inside the \texttt{<persona>} tags. Your task is to gradually communicate the content in \texttt{<instructions>} to the agent \ldots \\
Judge model & ``You are a strict evaluator of task execution results. You will be given: 1. task: the task description 2. init\_state: the initial state 3. final\_state: the final state 4. final\_assistant\_response: the assistant's \ldots \\
\end{tabularx}

\tcbline

\textbf{\color{ActionBlue}[4] Demonstration} \hfill \textit{message excerpt}\\[1pt]

\begin{minipage}[t]{\textwidth}
\begin{tcolorbox}[
  enhanced,
  colback=DemoUserBg,
  colframe=DemoUserFrame,
  boxrule=0.45pt,
  arc=1.2pt,
  colbacktitle=HeaderBlue,
  coltitle=RoleUser,
  fonttitle=\bfseries\scriptsize,
  left=3pt,right=3pt,top=1pt,bottom=1pt
]
\caseusertag\hspace{0.4em}
\textit{Initial request}\\[0pt]
Please review \texttt{LF-2024-PI-024}: first confirm the client, responsible attorney, and opposing party; then check whether the archived incident-investigation memorandum is still the current reliable version for linking; run a completeness review of the case file with special attention to the evidence chain; if the matter is stayed but not closed, add a new evidence item for store \ldots
\end{tcolorbox}
\end{minipage}

\vspace{3pt}

\begin{minipage}[t]{\textwidth}
\begin{tcolorbox}[
  enhanced,
  colback=DemoTraceBg,
  colframe=DemoTraceFrame,
  boxrule=0.45pt,
  arc=1.2pt,
  colbacktitle=ScaleGreen,
  coltitle=RoleAssistant,
  fonttitle=\bfseries\scriptsize,
  left=3pt,right=3pt,top=1pt,bottom=1pt
]
\caseassistanttag\hspace{0.4em}\textit{Plan}
Resolve the matter by number, verify party ownership, inspect \texttt{DOC-003} and its version history, validate integrity and traverse the evidence chain around \texttt{EVD-004}, then insert the new evidence item and re-check the post-update \ldots\\[3pt]
\casetooltag\hspace{0.4em}\textit{Tool call}
\texttt{validate\_matter\_integrity(MAT-2024-002)} $\rightarrow$ 1 violation; \texttt{EvidenceLink LNK-003 creates a self-loop}. \texttt{add\_evidence(..., evidence\_id=EVD-005)} $\rightarrow$ ID already exists; retry with \texttt{EVD-006} succeeds and links the new record to \texttt{DOC-003}.\\[3pt]
\caseassistanttag\hspace{0.4em}\textit{Final response}
The final workspace keeps the matter in \texttt{stayed} status, confirms \texttt{VER-004} as the current filed memo version, surfaces the self-loop anomaly on \texttt{LNK-003}, adds \texttt{EVD-006} with the requested metadata, and records the new filing \ldots

\end{tcolorbox}
\end{minipage}

{\color{MutedText}\textit{[\ldots additional turns omitted \ldots]}}



\tcbline

\textbf{\color{ActionBlue}[5] Rubrics and Judgment} \hfill \textit{selected rubric excerpts}\\[1pt]
\textbf{Overall pass@1: Pass ($\checkmark$)} \hspace{1.2em}
\textbf{Rubric: 20/20} \hspace{1.2em}
\textbf{Steps: 21}\\[2pt]
\begin{tabularx}{\textwidth}{@{}p{0.485\textwidth}@{\hspace{0.01\textwidth}}p{0.485\textwidth}@{}}
\begin{minipage}[t]{\linewidth}
\begin{tcolorbox}[
  colback=DemoTraceBg,
  colframe=DemoTraceFrame,
  boxrule=0.4pt,
  arc=1pt,
  title={\textbf{Rubric T1}\hfill\textbf{3/3 pts}},
  colbacktitle=ScaleGreen,
  coltitle=RoleAssistant,
  fonttitle=\scriptsize,
  left=3pt,right=5pt,top=3pt,bottom=3pt
]
\textit{<Criterion>}\\[1pt]
Perform an integrity review for the matter and detect the evidence-chain problem, specifically the self- \ldots \\[2pt]
\textit{<Judge reason>}\\[1pt]
The trajectory correctly resolves \texttt{LF-2024-PI-024}, confirms Yunhe Foods / Lin Qiaoxue / He Shan, verifies \texttt{VER-004} as current, and flags the self-referential \texttt{LNK-003} link.
\end{tcolorbox}
\end{minipage}
&
\begin{minipage}[t]{\linewidth}
\begin{tcolorbox}[
  colback=DemoTraceBg,
  colframe=DemoTraceFrame,
  boxrule=0.4pt,
  arc=1pt,
  title={\textbf{Rubric T4}\hfill\textbf{3/3 pts}},
  colbacktitle=ScaleGreen,
  coltitle=RoleAssistant,
  fonttitle=\scriptsize,
  left=3pt,right=5pt,top=3pt,bottom=3pt
]
\textit{<Criterion>}\\[1pt]
Create a corresponding matter event for the supplement and ensure the activity timeline reflects the addition on \ldots \\[2pt]
\textit{<Judge reason>}\\[1pt]
After one rejected ID attempt, the agent successfully adds \texttt{EVD-006}, links it to \texttt{DOC-003}, and confirms the resulting record through \texttt{event\_11}, the matter timeline, and the full \ldots.
\end{tcolorbox}
\end{minipage}
\end{tabularx}

\end{tcolorbox}

\subsection{Case 2}

\par\medskip
\noindent
\small
\begin{tcolorbox}[
  enhanced,
  breakable,
  colback=white,
  colframe=RuleGray,
  boxrule=0.6pt,
  arc=1.5pt,
  title={\textbf{Appendix Case Example: Expanded Litigation Matter Evidence Update}},
  colbacktitle=HeaderGray,
  coltitle=black,
  fonttitle=\bfseries\small,
  left=4pt, right=4pt, top=3pt, bottom=3pt
]
\scriptsize
\textbf{\color{ActionBlue}[1] Task Description} \hfill \textit{expanded task excerpt}\\[1pt]
Please help verify the Yunhe Foods personal-injury defense matter \texttt{LF-2024-PI-024}. The request proceeds in stages: first confirm the client, responsible attorney, and opposing party; then check whether the archived accident-investigation memorandum remains the current usable filed version; then run a completeness review of the matter with particular attention to evidence-chain anomalies. If the matter is stayed but not closed, and the memorandum can be used for attachment/linkage, add a newly collected evidence item titled ``store surveillance screenshots and incident timeline explanation'' with source \textit{store surveillance export}, collected on 2024-05-15, filed on 2024-05-16, authenticity pending verification, relevance high, custody secured, and a description tied to lighting, warning signage, and fall location, linked to \texttt{DOC-003}. Finally, reconfirm the trace record for this filing, inspect the full matter activity timeline, and return an updated workspace view that still exposes any unresolved completeness conflict.\\[2pt]

\tcbline

\textbf{\color{ActionBlue}[2] Tools} \hfill \textit{expanded tool summary}\\[1pt]
\texttt{get\_matter\_by\_number}, \texttt{get\_client\_by\_id}, \texttt{get\_attorney\_by\_id}, \texttt{get\_matter\_full\_context}, \texttt{get\_document\_by\_id}, \texttt{list\_document\_versions}, \texttt{validate\_matter\_integrity}, \texttt{get\_evidence\_chain\_subgraph}, \texttt{get\_evidence\_chain\_neighbors}, \texttt{get\_evidence\_by\_id}, \texttt{add\_evidence}, \texttt{list\_matter\_events}, and \texttt{get\_matter\_activity\_timeline}. These tools jointly expose matter identity, party records, current document-version state, graph-structured evidence relations, state-changing evidence insertion, and post-update audit trails.\\[2pt]

\tcbline

\textbf{\color{ActionBlue}[3] Agent Setup} \hfill \textit{prompt excerpts}\\[1pt]
\begin{tabularx}{\textwidth}{@{}>{\bfseries}p{0.17\textwidth}X@{}}
Tested agent & ``You are a helpful assistant. Your goal is to complete the user's request in an interactive environment by gradually calling the available tools step by step, and to proactively communicate with the user when necessary until the user ends the conversation. At each step, you will receive one of two types of information: a user reply, or a tool-call result returned by the environment.'' \\
User agent & ``You are role-playing as a user interacting with an agent. Your persona is written inside the \texttt{<persona>} tags. Your task is to gradually communicate the content in \texttt{<instructions>} to the agent through a multi-turn conversation.'' \\
Judge model & ``You are a strict evaluator of task execution results. You will be given: 1. task: the task description 2. init\_state: the initial state 3. final\_state: the final state 4. final\_assistant\_response: the assistant's final response to the user, if any 5. trajectory\_summary: a summary of the execution trajectory, including tool calls, parameters, and key observations 6. rubric\_items: the list of rubric items to judge one by one. Your goal: Based on this evidence, make a binary judgment for each rubric item.'' \\
\end{tabularx}

\tcbline

\textbf{\color{ActionBlue}[4] Demonstration} \hfill \textit{expanded message excerpt}\\[1pt]

\begin{minipage}[t]{\textwidth}
\begin{tcolorbox}[
  enhanced,
  colback=DemoUserBg,
  colframe=DemoUserFrame,
  boxrule=0.45pt,
  arc=1.2pt,
  colbacktitle=HeaderBlue,
  coltitle=RoleUser,
  fonttitle=\bfseries\scriptsize,
  left=3pt,right=3pt,top=1pt,bottom=1pt
]
\caseusertag\hspace{0.4em}
\textit{Initial request}\\[0pt]
Please review \texttt{LF-2024-PI-024}: confirm the client, responsible attorney, and opposing party; verify whether the archived incident-investigation memorandum is still the current filed version that can be relied upon for linking; run a completeness review of the full matter with special attention to evidence-chain anomalies; if the matter is stayed but not closed, add a new evidence item for store-surveillance screenshots and incident-timeline explanation, linked to \texttt{DOC-003}; then reconfirm the resulting trace record, inspect the full activity timeline, and return an updated workspace summary that still shows the unresolved integrity issue for later follow-up. 
\end{tcolorbox}
\end{minipage}

\vspace{3pt}

\begin{minipage}[t]{\textwidth}
\begin{tcolorbox}[
  enhanced,
  colback=DemoTraceBg,
  colframe=DemoTraceFrame,
  boxrule=0.45pt,
  arc=1.2pt,
  colbacktitle=ScaleGreen,
  coltitle=RoleAssistant,
  fonttitle=\bfseries\scriptsize,
  left=3pt,right=3pt,top=1pt,bottom=1pt
]
\caseassistanttag\hspace{0.4em}\textit{Plan}
Resolve the matter from \texttt{LF-2024-PI-024}, pull the linked client / attorney / party records, inspect \texttt{DOC-003} and its version history, run an integrity check over the matter graph, traverse the evidence neighborhood around \texttt{EVD-004}, then decide whether the stated preconditions for evidence insertion hold before creating the supplemental record and validating the resulting event / timeline / workspace views.\\[3pt]
\casetooltag\hspace{0.4em}\textit{Tool sequence}
\texttt{get\_matter\_by\_number(LF-2024-PI-024)} $\rightarrow$ \texttt{MAT-2024-002}, status \texttt{stayed}, attorney \texttt{ATT-002}, client \texttt{CLT-002}. 
\texttt{get\_client\_by\_id(CLT-002)} $\rightarrow$ Yunhe Foods Co. 
\texttt{get\_attorney\_by\_id(ATT-002)} $\rightarrow$ Lin Qiaoxue. 
\texttt{get\_matter\_full\_context(MAT-2024-002)} $\rightarrow$ opposing party He Shan, plaintiff-side counsel Qin Yiran. 
\texttt{get\_document\_by\_id(DOC-003)} and \texttt{list\_document\_versions(DOC-003)} $\rightarrow$ single filed current version \texttt{VER-004}.\\[3pt]
\casetooltag\hspace{0.4em}\textit{Integrity review}
\texttt{validate\_matter\_integrity(MAT-2024-002)} $\rightarrow$ one violation. 
\texttt{get\_evidence\_chain\_subgraph(seed=EVD-004, depth=3)} $\rightarrow$ only \texttt{EVD-004} plus link \texttt{LNK-003}. 
\texttt{get\_evidence\_chain\_neighbors(EVD-004)} $\rightarrow$ both inbound and outbound edges point back to \texttt{EVD-004}, confirming a self-referential loop explicitly marked as invalid.\\[3pt]
\casetooltag\hspace{0.4em}\textit{State change and verification}
\texttt{add\_evidence(..., evidence\_id=EVD-005)} $\rightarrow$ rejected because the identifier already exists. 
Retry with \texttt{EVD-006} $\rightarrow$ success; the new evidence is documentary, source \textit{store surveillance export}, collected on 2024-05-15, submitted on 2024-05-16, authenticity pending, relevance high, custody secured, linked to \texttt{DOC-003}. 
\texttt{get\_evidence\_by\_id(EVD-006)} confirms the inserted record. 
\texttt{list\_matter\_events(...)} returns \texttt{event\_11}; \texttt{get\_matter\_activity\_timeline(MAT-2024-002)} places the new filing after the 2024-05-14 stay and before the postponed 2024-05-16 status conference; \texttt{get\_matter\_full\_context(MAT-2024-002)} returns the updated workspace with both evidence records and the original unresolved self-loop.\\[3pt]
\caseassistanttag\hspace{0.4em}\textit{Final response}
The assistant returns a consolidated workspace summary: matter \texttt{LF-2024-PI-024} remains \texttt{stayed}; party ownership and adverse-side records are consistent; \texttt{DOC-003} remains anchored to current filed version \texttt{VER-004}; \texttt{EVD-006} is added successfully with the requested metadata and memo linkage; a new evidence-added event \texttt{event\_11} appears on 2024-05-16; and the prior completeness conflict persists as \texttt{LNK-003}, a self-loop on \texttt{EVD-004} that is preserved for later remediation rather than silently removed.

\end{tcolorbox}
\end{minipage}

\tcbline

\textbf{\color{ActionBlue}[5] Rubrics and Judgment} \hfill \textit{expanded rubric excerpts}\\[1pt]
\textbf{Overall pass@1: Pass ($\checkmark$)} \hspace{1.2em}
\textbf{Rubric: 20/20} \hspace{1.2em}
\textbf{Steps: 21}\\[2pt]
\begin{tabularx}{\textwidth}{@{}p{0.485\textwidth}@{\hspace{0.01\textwidth}}p{0.485\textwidth}@{}}
\begin{minipage}[t]{\linewidth}
\begin{tcolorbox}[
  colback=DemoTraceBg,
  colframe=DemoTraceFrame,
  boxrule=0.4pt,
  arc=1pt,
  title={\textbf{Rubric G4 / T1}\hfill\textbf{5/5 pts}},
  colbacktitle=ScaleGreen,
  coltitle=RoleAssistant,
  fonttitle=\scriptsize,
  left=3pt,right=5pt,top=3pt,bottom=3pt
]
\textit{<Criterion>}\\[1pt]
Check the current archival/document-version status for the incident-investigation memorandum, and perform an integrity review that detects the evidence-chain problem, specifically the self-referential link anomaly.\\[2pt]
\textit{<Judge reason>}\\[1pt]
The trajectory retrieves \texttt{DOC-003}, inspects the version list, confirms that \texttt{VER-004} is the sole filed current version, then runs the integrity validator and evidence-chain traversal tools, all of which consistently identify \texttt{LNK-003} as a self-loop on \texttt{EVD-004}.
\end{tcolorbox}
\end{minipage}
&
\begin{minipage}[t]{\linewidth}
\begin{tcolorbox}[
  colback=DemoTraceBg,
  colframe=DemoTraceFrame,
  boxrule=0.4pt,
  arc=1pt,
  title={\textbf{Rubric T2 / T3 / T4}\hfill\textbf{9/9 pts}},
  colbacktitle=ScaleGreen,
  coltitle=RoleAssistant,
  fonttitle=\scriptsize,
  left=3pt,right=5pt,top=3pt,bottom=3pt
]
\textit{<Criterion>}\\[1pt]
Add the new evidence item with the required metadata and memo linkage, ensure the description captures lighting / signage / fall-location support, and confirm that the supplement is reflected in the matter event log and activity timeline without changing the matter's stayed status.\\[2pt]
\textit{<Judge reason>}\\[1pt]
After one failed identifier attempt, the assistant retries with \texttt{EVD-006}, inserts the full evidence record, verifies the inserted fields directly, confirms the generated evidence-added event \texttt{event\_11} on 2024-05-16, and returns a post-update workspace in which the matter remains \texttt{stayed} while the new filing is visible in the timeline.
\end{tcolorbox}
\end{minipage}
\end{tabularx}

\end{tcolorbox}

\end{document}